\documentclass[10pt,twocolumn,letterpaper]{article}

\usepackage{cvpr}              

\usepackage[accsupp]{axessibility}
\usepackage{graphicx}
\usepackage{amsmath}
\usepackage{amssymb}
\usepackage{booktabs}

\usepackage[noend]{algpseudocode}
\usepackage{algorithmicx,algorithm}
\usepackage{bm}
\usepackage{pifont}
\usepackage{multirow}

\usepackage[pagebackref,breaklinks,colorlinks]{hyperref}

\usepackage[capitalize]{cleveref}
\crefname{section}{Sec.}{Secs.}
\Crefname{section}{Section}{Sections}
\Crefname{table}{Table}{Tables}
\crefname{table}{Tab.}{Tabs.}

\def\cvprPaperID{10304} 
\def\confName{CVPR}
\def\confYear{2022}

\begin{document}

\title{Overcoming Catastrophic Forgetting in Incremental Object Detection \\ via Elastic Response Distillation}

\author{Tao Feng$^1$, \quad Mang Wang$^{1}$\footnotemark[1] , \quad Hangjie Yuan$^2$\\
$^1$Alibaba Group \quad $^2$Zhejiang University\\
{\tt\small fengtao.hi@gmail.com, wangmang.wm@alibaba-inc.com, hj.yuan@zju.edu.cn}
}
\maketitle

\renewcommand{\thefootnote}{\fnsymbol{footnote}}
\footnotetext[1]{Corresponding author.}
\renewcommand{\thefootnote}{\arabic{footnote}}

\begin{abstract}
Traditional object detectors are ill-equipped for incremental learning. However, fine-tuning directly on a well-trained detection model with only new data will lead to catastrophic forgetting. Knowledge distillation is a flexible way to mitigate catastrophic forgetting. In Incremental Object Detection (IOD), previous work mainly focuses on distilling for the combination of features and responses. However, they under-explore the information that contains in responses. In this paper, we propose a response-based incremental distillation method, dubbed Elastic Response Distillation (ERD), which focuses on elastically learning responses from the classification head and the regression head. Firstly, our method transfers category knowledge while equipping student detector with the ability to retain localization information during incremental learning. In addition, we further evaluate the quality of all locations and provide valuable responses by the Elastic Response Selection (ERS) strategy. Finally, we elucidate that the knowledge from different responses should be assigned with different importance during incremental distillation. Extensive experiments conducted on MS COCO demonstrate our method achieves state-of-the-art result, which substantially narrows the performance gap towards full training. Code is available at \url{https://github.com/Hi-FT/ERD}.
\end{abstract}

\section{Introduction}

In the natural world, the visual system of creatures could constantly acquire, integrate and optimize knowledge. Learning mode is inherently incremental for them. In contrast, currently, the classic training paradigm of object detection models \cite{DBLP:conf/iccv/TianSCH19,DBLP:conf/cvpr/LiW0LT021} does not have such capability. Supervised object detection paradigm relies on accessing pre-defined labeled data. This learning paradigm implicitly assumes data distribution is fixed or stationary \cite{DBLP:journals/pr/FengJBLZ22, yuan2021DIN}, while data from real world is represented by continuous and dynamic data flow, whose distribution is non-stationary. When the model continuously obtains knowledge from non-stationary distribution, new knowledge would interfere with the old one, triggering catastrophic forgetting \cite{goodfellow2015empirical, 1989Catastrophic}. Based on whether the task identity is provided or must be inferred \cite{DBLP:journals/corr/abs-1904-07734}, researchers divide Incremental Learning (IL) into three types: task/domain/class IL. In this paper, we focus on the most intractable scenario for object detection: class incremental object detection.

A flexible way to solve IOD is knowledge distillation \cite{DBLP:journals/corr/HintonVD15}. \cite{DBLP:journals/cviu/PengZMLL21} stressed that the Tower layers could reduce catastrophic forgetting significantly. They implemented incremental learning on an anchor-free detector and selectively performed distillation on non-regression outputs. Meanwhile, in knowledge distillation for object detection where incremental learning was not introduced, previous work extracted knowledge from the combined distillation of different components. For example, \cite{DBLP:conf/nips/ChenCYHC17} and \cite{DBLP:journals/corr/abs-2006-13108} distilled all components of the detector. Nevertheless, the nature of these methods are designed using feature-based knowledge distillation \cite{DBLP:conf/cvpr/Chen0ZJ21}, response-based method \cite{DBLP:journals/ijcv/GouYMT21} has not been explored in IOD \cite{DBLP:journals/tnn/LiuKCXYZ21} yet. Besides, the advantage of response-based method is that it provides the reasoning information \cite{DBLP:journals/corr/HintonVD15,DBLP:conf/nips/MullerKH19} of the teacher detector. Therefore, an elaborate design for different responses is essential \cite{DBLP:conf/iccv/LinGGHD17}.

\begin{figure}
\centering
\includegraphics[width=0.35\textwidth]{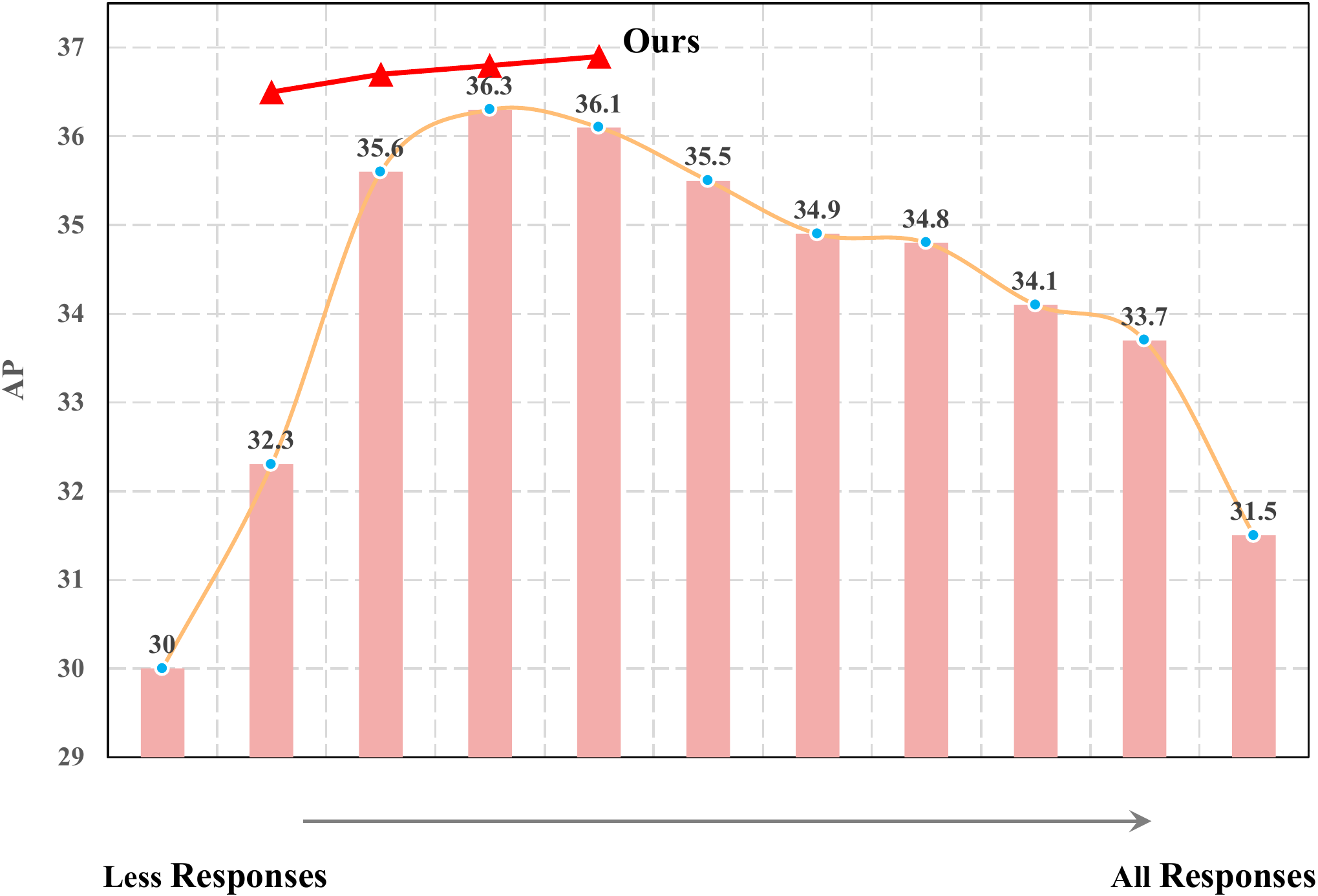}
\caption{The effect of various responses for IOD.}
\label{figure0}
\vspace{-1.3em}
\end{figure}

This paper focuses on a practical and challenging problem concerning IOD: \emph{how to learn response from classification predictions and bounding boxes}. Responses in object detection contain logits together with the offset of bounding box \cite{DBLP:journals/ijcv/GouYMT21}. Firstly, since the number of ground truth on each new image is uncertain, one of the foremost considerations is to validate the response of all samples, determining which response is positive or negative and which response each object should regress towards. Furthermore, as shown in Figure \ref{figure0}, we find that not all responses are important to prevent catastrophic forgetting, thus an appropriate number of response nodes is ideal. \cite{DBLP:journals/corr/KirkpatrickPRVD16} also proposed that synaptic consolidation achieves continuous learning by reducing synaptic plasticity critical to previous learning tasks. To sum up, we guide the student detector following the behavior of teacher on the old objects by constraining important responses to stay close to their old values.

To tackle the above problems, this paper rethinks response-based knowledge distillation method, finding that distillation at proper locations is crucial for facilitating IOD. Driven by this inspiration, we proposed an \textbf{E}lastic \textbf{R}esponse \textbf{D}istillation (ERD) scheme that elastically learns responses from classification head and regression head respectively. Unlike previous work, we introduce incremental localization distillation \cite{zheng2021localization} in regression response to equip student detector with the ability to learn location ambiguity \cite{DBLP:conf/nips/0041WW00LT020} during incremental learning. Besides, we propose \textbf{E}lastic \textbf{R}esponse \textbf{S}election (ERS) strategy to automatically select distillation nodes based on statistical characteristics from different responses, which evaluates the qualities of all locations and provides valuable responses. In this paper, we explain how we implement the constraint, and finally how we determine which responses are important. We greatly alleviate catastrophic forgetting problem and significantly narrow the gap with full training. Extensive experiments on the MS COCO dataset support our analysis and conclusion.

The our contributions can be summarized as follows,

\textbf{(i)} To the best of our knowledge, this paper is the first work to explore the response-based distillation method in IOD and dissect the essential differences between feature-based and response-based solutions for IOD.
\textbf{(ii)} We propose ERD based on statistical analysis, which separately distills selective classification and regression responses using the proposed ERS strategy.
\textbf{(iii)} Extensive experiments on MS COCO demonstrate that the proposed method achieves state-of-the-art performance and can be easily extended to different detectors.




\section{Related work}

\noindent\textbf{Incremental Learning.} Catastrophic forgetting is the core challenge for incremental learning. Incremental learning based on parameter constraints is a candidate solution for such problem, which protects the old knowledge by introducing an additional parameter-related regularization term to modify the gradient. EWC \cite{DBLP:journals/corr/KirkpatrickPRVD16} and MAS \cite{DBLP:conf/eccv/AljundiBERT18} are two typical representatives of such method. Another solution is incremental learning based on knowledge distillation. This kind of method mainly projects old knowledge by transferring knowledge in old tasks to new tasks through knowledge distillation. LwF \cite{DBLP:journals/pami/LiH18a} is the first method that introduces the concept of knowledge distillation into incremental learning, in the purpose of making predictions of the new model on new tasks similar to that of the old model and thereby protecting the old knowledge in the form of knowledge transfer. However, it would cause knowledge confusion when the correlation between new and old tasks is low. iCaRL \cite{DBLP:conf/cvpr/RebuffiKSL17} algorithm uses knowledge distillation to avoid excessive deterioration of knowledge in the network, while BiC \cite{DBLP:conf/cvpr/WuCWYLGF19} added a bias correction layer after the FC layer to offset the category bias of new data when using the distillation loss.

\noindent\textbf{Incremental Object Detection.} Compared with incremental classification, IOD is less explored. Meanwhile, the high complexity of the detection task also adds the difficulty of incremental object detection. \cite{DBLP:conf/iccv/ShmelkovSA17} proposed to apply LwF to Fast RCNN detector \cite{girshick2015fast}, which is the first work on incremental object detection. Thereafter, some researchers move this area forward. \cite{DBLP:journals/cviu/PengZMLL21} proposed SID approach for IOD on anchor-free detector and conducted experiments on FCOS \cite{DBLP:conf/iccv/TianSCH19} and CenterNet \cite{zhou2019objects}. \cite{li2021classincremental} studied object detection based on class-incremental learning on Faster RCNN detector with emphasis on few-shot scenarios, which is also the focus of ONCE algorithm \cite{DBLP:conf/cvpr/Perez-RuaZHX20}. \cite{DBLP:conf/edge/LiTGZZH19} designed an incremental object detection system with RetinaNet detector \cite{DBLP:journals/pami/LinGGHD20} on edge devices. the latest work, \cite{DBLP:conf/cvpr/Joseph0KB21} introduced the concept of incremental learning when defining the problems of Open World Object Detection (OWOD).  However, existing IOD distillation framework does not pay enough attention to the significant role of head. In this study, we found head has its great significance in the area of IOD.

\begin{figure*}
\centering
\includegraphics[width=0.85\textwidth]{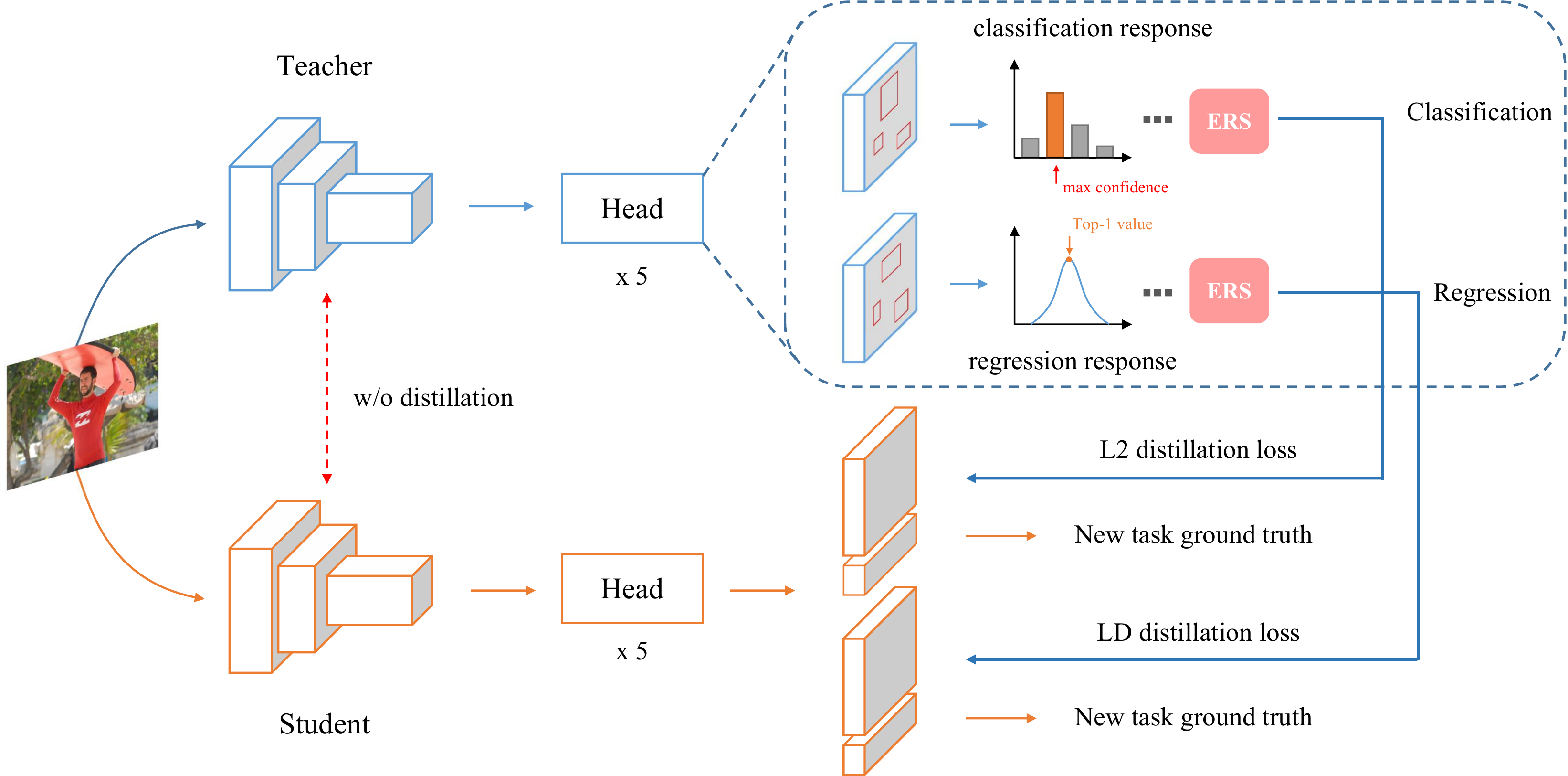}
\caption{Overall structure of elastic response distillation for incremental object detection.}
\label{figure1}
\vspace{-1.4em}
\end{figure*}

\noindent\textbf{Knowledge Distillation for Object Detection.} Knowledge distillation \cite{DBLP:conf/kdd/BucilaCN06, DBLP:conf/aaai/ChenMWF020} is an effective way to transfer knowledge between models. Widely applied in image classification tasks in previous researches, knowledge distillation is now used in object detection tasks frequently \cite{DBLP:conf/cvpr/DaiJWBW0Z21}. \cite{DBLP:conf/nips/ChenCYHC17} implemented distillation for all components of Faster RCNN (including backbone, proposals in RPN, and head). To imitate the high-level feature response of the teacher model with the student model, \cite{DBLP:conf/cvpr/WangYZF19} proposed a distillation method based on fine-grained feature imitation. By synthesize category-conditioned objects through inverse mapping, \cite{DBLP:conf/wacv/ChawlaYMA21} proposed a data-free distillation technology applicable for object detection, but the method would trigger dream-image. \cite{DBLP:conf/cvpr/Guo00W0X021} believing that foreground and background both play an unique role in object detection, proposed an object detection distillation method that decoupled foreground and background. \cite{zheng2021localization} proposed a localization distillation method introducing knowledge distillation into the regression branch of the detector, so as to enable the student network to solve the localization ambiguity as the teacher network.

\section{Method}

\subsection{Motivation}

The purpose of IOD is to transfer old knowledge to student detector, and this knowledge could be the features of intermediate layers in  backbone or neck, or the soft targets in head. Unlike feature-based method, response-based method can provides the reasoning information of teacher detector \cite{DBLP:journals/corr/HintonVD15,DBLP:conf/nips/MullerKH19}. Therefore, we incrementally learn a strong and efficient student object detector by the distillation of incremental knowledge from responses of different heads. 

\subsection{Overall Structure}

The overall framework of the proposed method is shown in Figure \ref{figure1}. Firstly, ERD is applied to learn elastic response from the classification head and regression head of the teacher detector. Secondly, incremental localization distillation loss is applied to enhance the localization information extraction ability of the student detector. Notably, the ERS strategies are proposed to gain more meaningful incremental responses from the teacher detector, that is, selective calculation of the distillation loss from the response provided by the teacher detector. The overall learning target of the student detector is therefore defined as,
\begin{equation}
\begin{aligned}
\mathcal{L}_{{total}}=\mathcal{L}_{model}&+\lambda_{1} \mathcal{L}_{ERD\_{cls}}(\mathcal{{C}_{T}}, \mathcal{{C}_{S}})
\\&+\lambda_{2} \mathcal{L}_{ERD\_{bbox}}(\mathcal{{B}_{T}}, \mathcal{{B}_{S}})
\end{aligned}    
\end{equation}
\noindent where $\lambda_{i}$ is the parameters that balances the weights of different loss terms, and the subscript $\mathcal{T}$ and $\mathcal{S}$ separately represents teacher and student. The loss term $\mathcal{L}_{model}$ is the detector-specific classification and localization loss to train student detector for detecting new objects. The second loss term $\mathcal{L}_{ERD\_{cls}}$ is the incremental L2 distillation loss for the classification branch. The third loss term $\mathcal{L}_{ERD\_{bbox}}$ is the incremental localization distillation loss for the regression branch. Both $\mathcal{L}_{ERD\_{cls}}$ and $\mathcal{L}_{ERD\_{bbox}}$ are used for the outputs of old classes. We use $\lambda_{1}=\lambda_{2}=1$ by default.

In the following subsection, we mainly present ERD and ERS for GFLV1 \cite{DBLP:conf/nips/0041WW00LT020} while we generalize our method to FCOS in Table \ref{table7}, which illustrates the effectiveness of our method. 

\subsection{ERD at Classification Head}

The soft predictions from the classification head contains the knowledge of various categories discovered by the teacher detector. Through the learning of soft predictions, the student model can inherit hidden knowledge, which is intuitive for classification tasks \cite{DBLP:journals/corr/HintonVD15}.  Let $\mathcal{T}$ be the teacher model, we use SoftMax to transform logits $\mathcal{{C}_{T}}$ into distribution, then the outputting probability distribution $\mathcal{{P}_{T}}$ is defined as,
\begin{equation}
\mathcal{{P}_{T}}=\operatorname{SoftMax}\left(\mathcal{{C}_{T}}/t\right)
\end{equation}
\noindent Similarly, we define $\mathcal{{P}_{S}}$ for the student model $\mathcal{S}$ as $\mathcal{{P}_{S}}=\operatorname{SoftMax}\left(\mathcal{{C}_{S}}/t\right)$, where $t$ is the temperature factor to soften the probability distribution for $\mathcal{{P}_{T}}$ and $\mathcal{{P}_{S}}$.

Previous work usually directly utilizes all the predicted responses in classification head and treat each position equally, e.g. $\mathcal{L}_{cls}=\sum_{i=1}^{N}\mathcal{L}_{K L}\left(\mathcal{{P}_{T}}, \mathcal{{P}_{S}}\right)$. If there is any inappropriate balance, the response generated by the background category may overwhelm the response generated by the foreground category, thereby interfering with the retention of old knowledge. Here, we selectively calculate the distillation loss from response, thus the incremental distillation loss at classification head is as follows,
\begin{equation}
\mathcal{L}_{{ERD\_{cls}}}\left(\mathcal{{C}_{T}}, \mathcal{{C}_{S}}\right)=\sum_{i=1}^{m}\left(\mathcal{{C}_{T}}^{i}-\mathcal{{C}_{S}}^{i}\right)^{2}
\label{eq4}
\end{equation}
\noindent where $\mathcal{{C}_{T}}^{i}$  is one of the $m$ selected category responses from the teacher detector using the new data. $\mathcal{{C}_{S}}^{i}$ is the corresponding category responses of the student detector. By distilling the selected responses, the student detector incrementally inherits the old knowledge of the teacher detector.

\subsection{ERD at Regression Head}

The bounding box responses from the regression branch are also important for IOD. Contrary to the discrete class information, the output of regression branch may provide a regression direction contradicting the real direction. Even if an image does not contain any objects of old categories, the regression branch would still predict bounding boxes, though the confidence is low. This poses a challenge for transferring regression knowledge from teacher detector to student detector. Furthermore, in previous work, only the bounding boxes of objects with high classification confidence are utilized as the regression knowledge from the teacher detector, which ignores the localization information of regression branch.

Benefitting from the general representation of distributions for bounding boxes from GFLV1 detector, each edge $e$ of a bounding box can be represented as a probability distribution through SoftMax function \cite{zheng2021localization}. Thus, the probability matrix of each bounding box $\mathcal{B}$ can be defined as, 
\begin{equation}
\mathcal{B}=\left[p_{t}, p_{b}, p_{l}, p_{r}\right] \in \mathbb{R}^{n \times 4}
\end{equation}

Therefore, we can extract the incremental localization knowledge of bounding box $\mathcal{B}$ from the teacher detector $\mathcal{T}$ and transfer it to the student detector $\mathcal{S}$ using KL-Divergence loss, 
\begin{equation}
\mathcal{L}_{L D}^{j}= \sum_{e \in \mathcal{B}} \mathcal{L}_{K L}^{e}\left(\mathcal{{B}_{T}}^{j}, \mathcal{{B}_{S}}^{j}\right)
\end{equation}

Finally, the incremental localization distillation loss at regression head is defined as,
\begin{equation}
\mathcal{L}_{{ERD\_{bbox}}}\left(\mathcal{{B}_{T}}, \mathcal{{B}_{S}}\right)=\sum_{j=1}^{J} \mathcal{L}_{L D}^{j}
\label{eq7}
\end{equation}
\noindent where $\mathcal{{B}_{T}}^{j}$ is the regression response of the teacher detector from $J$ selected bounding boxes using the new data, and $\mathcal{{B}_{S}}^{j}$ is the corresponding regression response of the student detector. Notably, the incremental localization distillation provides extra localization information.

\subsection{Elastic Response Selection}

As shown in Figure \ref{figure0}, choosing all the responses leads to bad performance, thus response selection is important to prevent catastrophic forgetting. Then a natural question arises: \emph{how to select responses as the distillation nodes}. Common selection strategies depend on sensitive hyper-parameters such as setting confidence thresholds or selecting Top-K scores. These empirical practices may result in a consequence that small thresholds ignore several old objects while large ones bring negative responses.

To solve the above problem, we propose the ERS strategy as illustrated in Algorithm 1. We respectively select responses from the classification head and regression head as the distillation nodes.



\noindent \textbf{Classification head.} Statistical characteristics are utilized to select responses of the classification head, as described in L-3 to L-11. Specifically, We first calculate the confidence score of each node. After that, we calculate the mean $\mu_{C}^{\prime}$ and standard deviation $\sigma_{C}^{\prime}$ in L-5 and L-6. With these statistics, the elastic threshold $\tau_{C}^{\prime}$ can be obtained in L-7. Finally, we select response nodes whose confidence scores are greater than the threshold $\tau_{C}^{\prime}$ in L-8 to L-11 as the distillation nodes.

\noindent \textbf{Regression head.} Statistical distribution information is utilized to select responses of the regression head, as described in L-13 to L-22. For GFLV1, a certain and unambiguous bounding box usually has a sharper distribution. Therefore, the Top-1 value is relatively larger if the distribution is sharp. Based on the above statistical properties, the Top-1 value is used to measure the confidence of each bounding box. Specifically, we first select the Top-1 value of each distribution. After that, we calculate the mean $\mu_{B}^{\prime}$ and the standard deviation $\sigma_{B}^{\prime}$ of all Top-1 values in L-15 and L-16. Then, the threshold  $\tau_{B}^{\prime}$ is obtained in L-17. Finally, we select these candidates whose confidence are greater than the threshold  $\tau_{B}^{\prime}$ in L-18 to L-20. The \emph{nms} operator returns a sampled set that is filtered by NMS in L-21.

\begin{table*}
\small
\centering
\caption{Incremental results (\%) based on GFLV1 detector on COCO benchmark under different scenarios. (``$\Delta$'' represents an improvement over Catastrophic Forgetting. ``$\nabla$'' represents the gap towards the Upper Bound.)}
\label{table11}
\begin{tabular}{@{}l|l|cccccc@{}}
\toprule
Scenarios    & Method & $AP$ & $AP_{50}$ & $AP_{75}$  & $AP_{S}$  & $AP_{M}$  & $AP_{L}$  \\ \midrule
Full data    & Upper Bound & 40.2 & 58.3  & 43.6  & 23.2  & 44.1   & 52.2  \\ \midrule
    & Catastrophic Forgetting & 17.8 & 25.9  & 19.3 & 8.3  & 19.2 & 24.6 \\ \cmidrule(l){2-8}
\multirow{3}{*}{40 classes + 40 classes}    & LwF \cite{DBLP:journals/pami/LiH18a} & 17.2 ($\Delta -0.6/\nabla 23.0$) & 25.4  & 18.6 & 7.9  & 18.4 & 24.3\\
    & RILOD \cite{DBLP:conf/edge/LiTGZZH19} & 29.9 ($\Delta 12.1/\nabla 10.3$) & 45.0  & 32.0 & 15.8  & 33.0 & 40.5\\
    & SID \cite{DBLP:journals/cviu/PengZMLL21} & 34.0 ($\Delta 16.2/\nabla 6.2$) & 51.4  & 36.3 & 18.4  & 38.4 & 44.9\\
    & ERD  & \textbf{36.9 ($\Delta 19.1/\nabla 3.3$)} & \textbf{54.5}  & \textbf{39.6}  & \textbf{21.3}  & \textbf{40.4}   & \textbf{47.5}\\ \bottomrule
\toprule
    & Catastrophic Forgetting & 14.1 & 20.6  & 15.2 & 7.0  & 14.5 & 19.2\\ \cmidrule(l){2-8}
\multirow{3}{*}{50 classes + 30 classes}    & LwF \cite{DBLP:journals/pami/LiH18a} & 5.0 ($\Delta -9.1/\nabla 35.2$) & 9.5  & 4.6 & 5.0  & 6.7 & 5.7\\
    & RILOD \cite{DBLP:conf/edge/LiTGZZH19} & 28.5 ($\Delta 14.4/\nabla 11.7$) & 43.2  & 30.2 & 15.4  & 31.6 & 38.0\\
    & SID \cite{DBLP:journals/cviu/PengZMLL21} & 33.8 ($\Delta 19.7/\nabla 6.4$) & 51.0  & 36.1 & 17.6  & 38.1 & 45.1\\
    & ERD  & \textbf{36.6 ($\Delta 22.5/\nabla 3.6$)} & \textbf{54.0}  & \textbf{38.9}  & \textbf{19.4}  & \textbf{40.4}   & \textbf{48.0}\\ \bottomrule
\toprule
    & Catastrophic Forgetting & 9.8 & 14.0  & 10.6 & 4.3  & 14.1 & 13.5\\ \cmidrule(l){2-8}
\multirow{3}{*}{60 classes + 20 classes}    & LwF \cite{DBLP:journals/pami/LiH18a} & 5.8 ($\Delta -4.0/\nabla 34.4$) & 10.8  & 5.3 & 4.0  & 8.5 & 7.7\\
    & RILOD \cite{DBLP:conf/edge/LiTGZZH19} & 25.4 ($\Delta 15.6/\nabla 14.8$) & 38.8  & 26.8 & 13.9  & 29.0 & 33.7\\
    & SID \cite{DBLP:journals/cviu/PengZMLL21} & 32.7 ($\Delta 22.9/\nabla 7.5$) & 49.8  & 34.6 & 17.2  & 37.6 & 43.5\\
    & ERD  & \textbf{35.8 ($\Delta 26.0/\nabla 4.4$)} & \textbf{52.9}  & \textbf{38.4}  & \textbf{20.6}  & \textbf{39.4}   & \textbf{46.5}\\ \bottomrule
    \toprule
    & Catastrophic Forgetting & 4.3 & 6.5  & 4.5 & 2.1  & 5.1 & 6.8\\ \cmidrule(l){2-8}
\multirow{3}{*}{70 classes + 10 classes}    & LwF \cite{DBLP:journals/pami/LiH18a} & 7.1 ($\Delta 2.8/\nabla 33.1$) & 12.4  & 7.0 & 4.8  & 9.5 & 10.0\\
    & RILOD \cite{DBLP:conf/edge/LiTGZZH19} & 24.5 ($\Delta 20.2/\nabla 15.7$) & 37.9  & 25.7 & 14.2  & 27.4 & 33.5\\
    & SID \cite{DBLP:journals/cviu/PengZMLL21} & 32.8 ($\Delta 28.5/\nabla 7.4$) & 49.0  & 35.0 & 17.1  & 36.9 & 44.5\\
    & ERD  & \textbf{34.9 ($\Delta 30.6/\nabla 5.3$)} & \textbf{51.9}  & \textbf{37.4}  & \textbf{18.7}  & \textbf{38.8}   & \textbf{45.5}\\ \bottomrule
\end{tabular}
\vspace{-0.8em}
\end{table*}

The motivations behind ERS are explained as follows:

\noindent \textbf{Maintain fairness among different responses.} In a normal distribution, approximately 16\% and 2.5\% of the samples are separately distributed in the interval $[\mu+\sigma, +\infty]$ and $[\mu +2\sigma, +\infty]$. In our case, the number of positive responses are distributed from 100 to 1000 per image. In contrast, the strategy of selecting all or top-k responses leads to unfairness for different responses.

\noindent \textbf{Elastic selection by statistical characteristics.} In the IOD task, responses generated by background objects may overwhelm the responses generated by foreground objects. Thus a high $\mu$ indicates high-quality candidates, while a low one indicates low-quality candidates. ERS can elastically select enough positive responses following the statistical characteristics of different branches.

\section{Experiments and Discussions}

In this section, we perform experiments on MS COCO 2017 \cite{DBLP:journals/corr/ChenFLVGDZ15} using the baseline detector GFLV1 to validate our method. Then, we perform ablation studies to prove the effectiveness of each component. Finally, we discuss the application scenario of our method.

\noindent \textbf{Implementation Details.} We build our method on top of the GFLV1 detector. The teacher and student detectors defined in our experiments are standard GFLV1 architectures. For GFLV1 detector, ResNet-50 is used as its backbone, FPN \cite{DBLP:conf/cvpr/LinDGHHB17} is used as its neck. We train the detector to following the same settings as the original paper. All the experiments are performed on 8 NVIDIA Tesla V100 GPUs, with a batch size of 8. For the parameter $\alpha$, we use $\alpha_{1}=\alpha_{2}=2$ by default.

\noindent \textbf{Datasets and Evaluation Metric.} MS COCO 2017 is a challenging dataset in object detection which contains 80 object classes. For experiments on this dataset, we use the train set for training and the minival set for testing. The standard COCO protocols are used as the evaluation metrics, i.e. $AP$, $AP_{50}$, $AP_{75}$, $AP_{S}$, $AP_{M}$ and $AP_{L}$. 

\noindent\textbf{Experiment Setup.} The detector is trained by 12 epochs (1x mode) for each incremental step. The settings are consistent for all the detectors in the different scenarios. Specifically, we conduct experiments in the following Class Incremental Learning scenarios with different splits: 

\textbf{(i)} One-step: 40 + 40 to 70 + 10 with a step size of 10 classes, increasing base class numbers and decreasing new class numbers. \textbf{(ii)} Multi-step: two-step and four-step settings with  20 new classes and 10 new classes respectively added each time. \textbf{(iii)} Last 40 + First 40: last 40 classes as the base classes and first 40 classes as new classes.

\begin{figure*}
    \centering
   \begin{subfigure}{0.3\textwidth}
       \includegraphics[width=1.7in]{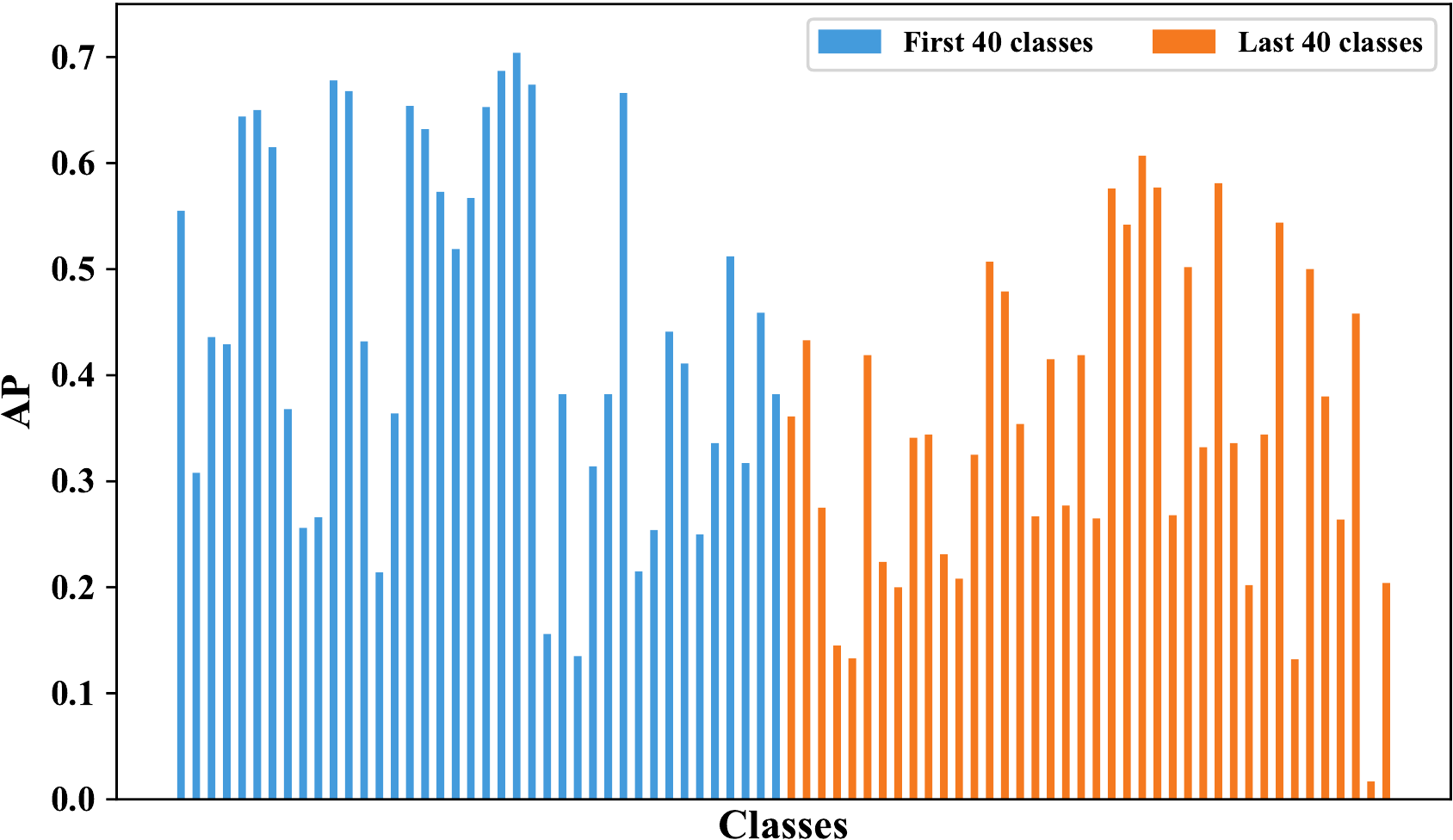}
        \caption{Upper Bound.}
        \label{fig:subfigure21}
    \end{subfigure}
   \begin{subfigure}{0.3\textwidth}
       \includegraphics[width=1.7in]{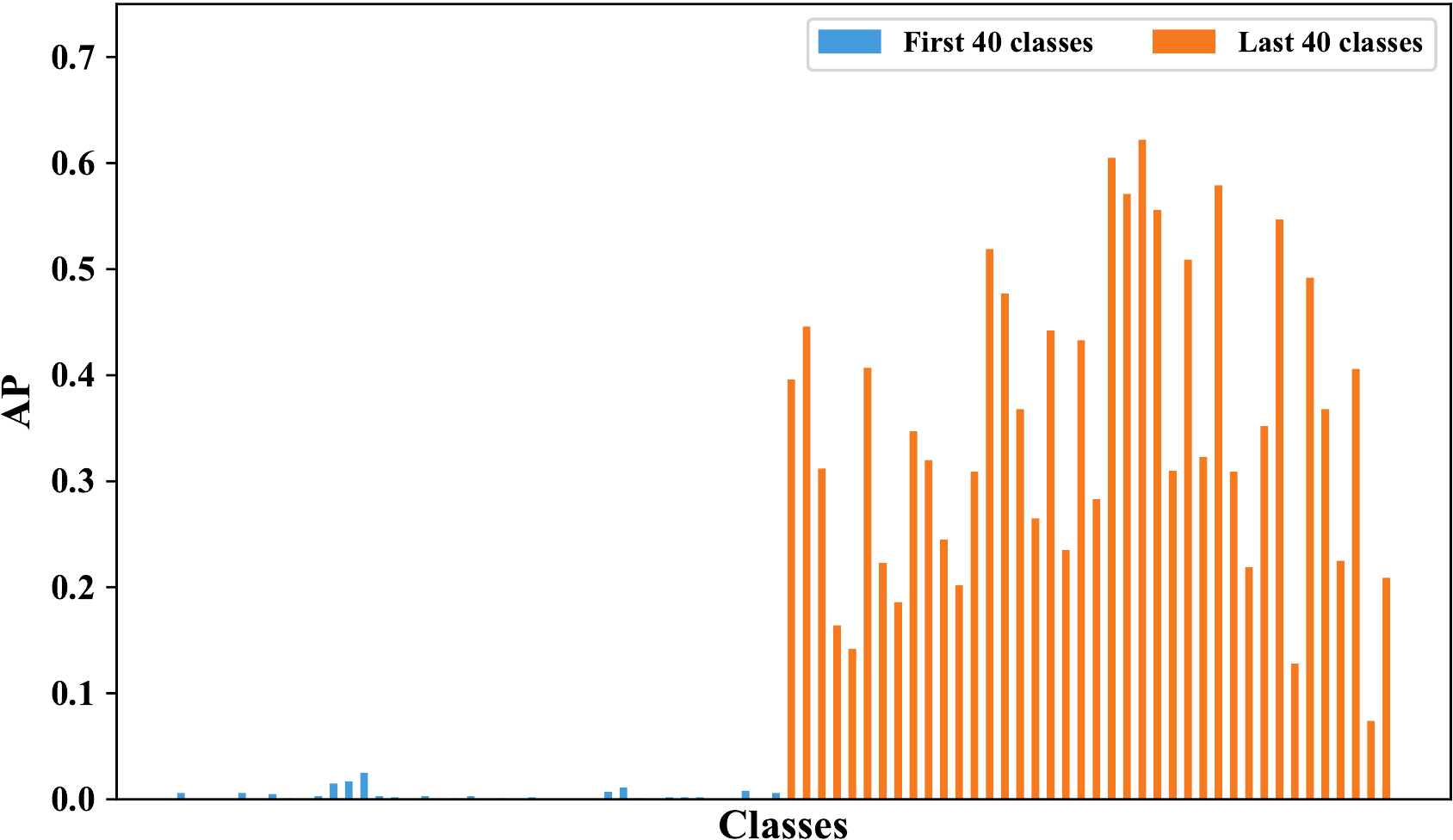}
        \caption{Catastrophic Forgetting.}
        \label{fig:subfigure22}
    \end{subfigure}
   \begin{subfigure}{0.3\textwidth}
       \includegraphics[width=1.7in]{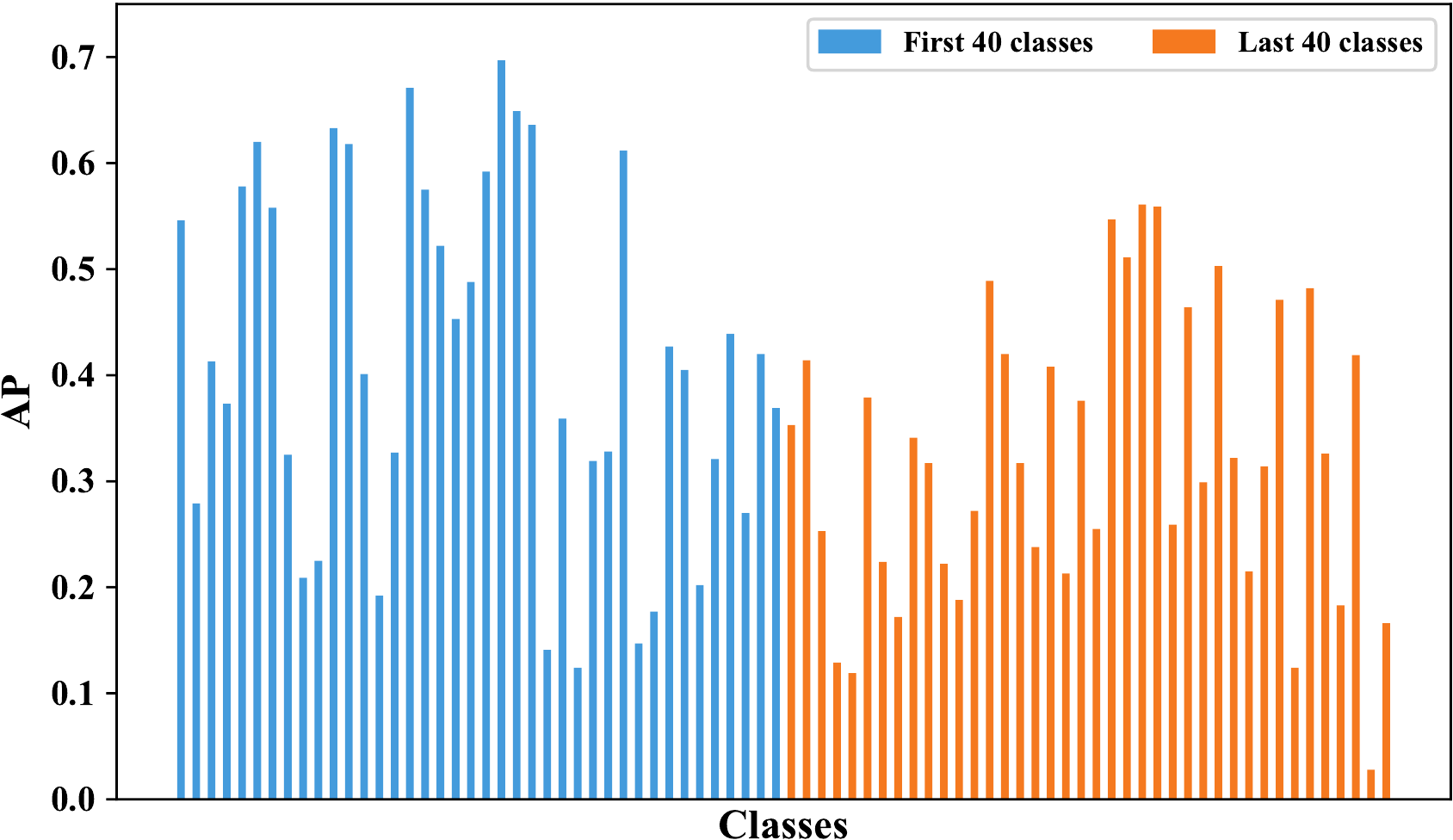}
        \caption{Elastic Response Distillation.}
        \label{fig:subfigure23}
    \end{subfigure}
\caption{AP of per-class among different learning schemes. (a) Detector is trained with all data.(b) Student detector is fine-tuned with new classes.(c) Student detector is learned via ERD.}
\label{figure2}
\vspace{-0.3em}
\end{figure*}

\begin{table*}
\centering
\caption{Incremental results ($AP/AP_{50}$, \%) based on GFLV1 detector on COCO benchmark under the four-step setting. A(a-b) is the one-step normal training for categories a-b and +B(c-d) is the incremental training for categories c-d.}
\label{table_four}
\begin{tabular}{@{}l|c|cccc|c@{}}
\toprule
Method    & A(1-40) & +B(40-50) & +B(50-60) & +B(60-70)  & +B(70-80) & A(1-80) \\ \midrule
Catastrophic Forgetting  & & 5.8/ 8.5  & 5.7/ 8.3 & 6.3/ 8.5 & 3.3/ 4.8  \\ 
RILOD \cite{DBLP:conf/edge/LiTGZZH19}    &\multirow{2}{1.5cm}{45.7/ 66.3 } & 25.4/ 38.9  & 11.2/ 17.3 & 10.5/ 15.6 & 8.4/ 12.5 & \multirow{2}{1.5cm}{40.2/ 58.3 }  \\
SID \cite{DBLP:journals/cviu/PengZMLL21}     & & 34.6/ 52.1  & 24.1/ 38.0 & 14.6/ 23.0 & 12.6/ 23.3 \\
ERD      & & \textbf{36.4/ 53.9}  & \textbf{30.8/ 46.7}  & \textbf{26.2/ 39.9}  & \textbf{20.7/ 31.8}  \\\bottomrule
\end{tabular}
\vspace{-0.6em}
\end{table*}

\begin{table}
\scriptsize
\centering
\caption{ Incremental results ($AP/AP_{50}$, \%) based on GFLV1 detector on COCO benchmark under the two-step setting, where the meanings of A(a-b) and +B(c-d) are similar to Table \ref{table_four}. }
\label{table_two}
\begin{tabular}{@{}l|c|cc|c@{}}
\toprule
Method    & A(1-40) & +B(40-60) & +B(60-80) & A(1-80)  \\ \midrule
Catastrophic Forgetting  &         & 10.7/ 15.8  & 9.4/ 13.3 &  \\ 
RILOD \cite{DBLP:conf/edge/LiTGZZH19}     & \multirow{2}{1.1cm}{45.7/ 66.3 }  & 27.8/ 42.8  & 15.8/ 4.0   & \multirow{2}{1.1cm}{40.2/ 58.3 } \\
SID \cite{DBLP:journals/cviu/PengZMLL21}  &         & 34.0/ 51.8  & 23.8/ 36.5 &  \\
ERD       &         & \textbf{36.7/ 54.6}  & \textbf{32.4/ 48.6} &  \\\bottomrule
\end{tabular}
\vspace{-1.8em}
\end{table}

\begin{table*}
\centering
\caption{Ablation study (\%) based on GFLV1 detector using the COCO benchmark under first 40 classes + last 40 classes. (``$\Delta$'' represents an improvement over Catastrophic Forgetting. ``$\nabla$'' represents the gap towards the Upper Bound.)}
\label{table3}
\begin{tabular}{@{}l|cccccc@{}}
\toprule
Method & $AP$ & $AP_{50}$ & $AP_{75}$  & $AP_{S}$  & $AP_{M}$  & $AP_{L}$  \\ \midrule
Upper Bound & 40.2 & 58.3  & 43.6  & 23.2  & 44.1   & 52.2  \\
Catastrophic Forgetting & 17.8 & 25.9  & 19.3 & 8.3  & 19.2 & 24.6\\ \midrule
KD:all cls + all reg & 31.5($\Delta 13.7/\nabla 8.7$) & 48.3  & 33.4 & 17.7  & 35.3 & 41.3\\
KD:all cls & 23.8($\Delta 6.0/\nabla 16.4$) & 36.6  & 24.9 & 11.8  & 27.2 & 32.9\\
KD:all reg & 13.0($\Delta -4.8 /\nabla 27.2$) & 21.1  & 13.4 & 5.0  & 14.7 & 18.6\\
ERD:cls + ERS & 33.2($\Delta 15.4/\nabla 7.0$) & 51.2  & 35.2 & 18.5  & 37.8 & 43.8\\
ERD:cls + reg + ERS & \textbf{36.9($\Delta 19.1/\nabla 3.3$)} & \textbf{54.5}  & \textbf{39.6}  & \textbf{21.3}  & \textbf{40.4}   & \textbf{47.5}\\ \bottomrule
\end{tabular}
\vspace{-0.4em}
\end{table*}

\subsection{Overall Performance}

\textbf{One-step.} We reported the incremental results under the first 40 classes + last 40 classes scenario in Table \ref{table11}. In this case, we observe that if the old detector and the new data are directly used to conduct fine-tuning process, then the AP drops to 17.8\% as compared to the 40.2\% in full data training (Upper Bound). This is because the fine-tuning process makes the detector's memory of old objects close to 0, resulting in catastrophic forgetting (ref to Figure \ref{fig:subfigure22}). Our method far outperformed fine-tuning across various evaluation criteria. Concretely, when the IoU is 0.5, 0.75 and 0.95, the AP respectively improves by 19.1\%, 28.6\% and 20.3\%, which indicates that our method can well address the catastrophic forgetting problem. Notably, even compared with the upper bound where the entire dataset is used for training, our method only has a performance gap of 3.3\%. It indicates that the student detector maintains a good memory of old objects while is able to learn knowledge of new objects. Remarkably, as shown in Table \ref{table11}, the performance of fine-tuning decreases drastically as the number of new classes decreases (17.8\% to 4.3\%) under different incremental conditions (50 classes + 30 classes, 60 classes + 20 classes, and 70 classes + 10 classes), while our method still remains a high level performance (36.9\% to 34.9\%). To sum up, our method has great robustness for overcoming catastrophic forgetting.

In addition, we compare our method with LwF \cite{DBLP:journals/pami/LiH18a}, RILOD \cite{DBLP:conf/edge/LiTGZZH19} and SID \cite{DBLP:journals/cviu/PengZMLL21} as well. Table \ref{table11} shows that although LwF works well in incremental classification, it has even lower AP than direct fine-tuning in detection task, which reveals naively borrowing methods from incremental classification area would generate negative influence to the IOD task. For the typical IOD approaches (i.e. RILOD and SID), in order to fairly compare with them, we replicate them based on the GFLV1 detector. For RILOD, we completely follow their implementations. For SID, we use the component with the greatest improvement in the paper. When compared with the aforementioned approaches, the proposed method achieves state-of-the-art performance in four incremental scenarios. Notably, the performance improvements are all significant.

\begin{figure}[htbp]
\centering
\includegraphics[width=4.9cm]{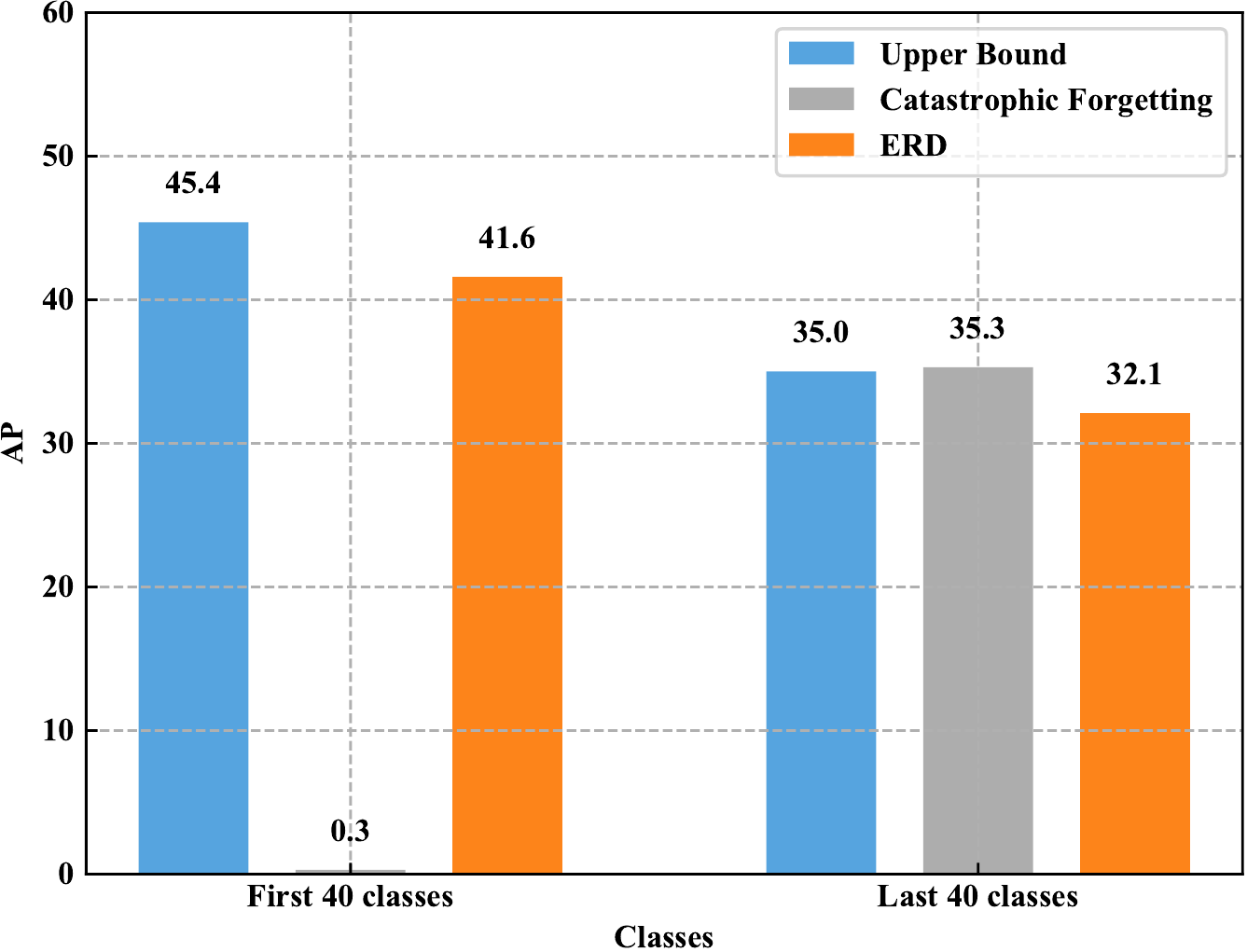}
\caption{AP of all classes in first 40 classes vs. last 40 classes.}
\label{figure4}
\vspace{-1.5em}
\end{figure}

To put it more intuitively, we visualize the AP of all classes in first 40 classes and last 40 classes in Figure \ref{figure4}. Furthermore, the per-class results are visualized in Figure \ref{figure2}, where the blue columns denote the per-class AP in first 40 classes, and the orange columns denote the per-class AP in last 40 classes. As Figure \ref{figure2} shows, the proposed method reserves a majority of information for the old classes while learns knowledge from newly coming classes.

\textbf{Multi-step.} We reported the incremental results under multi-step settings to illustrate the continual learning ability of the proposed method. In Table \ref{table_two} (two-step) and Table \ref{table_four} (four-step), our method outperforms fine-tuning by a large margin for each incremental step on both multi-step settings. This is because, the detector continuously obtains knowledge from the dynamic data flow, new knowledge interferes with the old one, triggering catastrophic forgetting, while ERD provides valuable responses in each step to alleviate the problem. In addition, ERD performs favorably well on each incremental step against the previous state-of-the-art. Remarkably, the $AP$ of RILOD and SID decreases drastically as the number of new classes increases (27.8\% to 15.8\% and 34.0\% to 23.8\%, 25.4\% to 8.4\% and 34.6\% to 12.6\%) under two multi-step settings, while our method still remains a high performance (36.7\% to 32.4\% and 36.4\% to 20.7\%). ERD is able to restore the previous class performance to a respectable level. It indicates that the proposed method has a significant ability to alleviate catastrophic forgetting.


\begin{table}
\small
\centering
\caption{Varying $\alpha$ for ERS (\%).}
\label{table5}
\begin{tabular}{@{}l|cccccc@{}}
\toprule
Threshold & $AP$ & $AP_{50}$ & $AP_{75}$ & $AP_{S}$  & $AP_{M}$  & $AP_{L}$  \\ \midrule
$\alpha_{1,2}=1,1$ & 36.5 & 54.2  & 39.2 & 20.6 & 40.3  & 46.9 \\ 
$\alpha_{1,2}=1,2$ & 36.8  & 54.4  & 39.6 & 21.5 & 40.4  & 47.5 \\
$\alpha_{1,2}=2,1$  & 36.7  & 54.3  & 39.6 & 21.5 & 40.4  & 47.6 \\
$\alpha_{1,2}=2,2$  & \textbf{36.9} & \textbf{54.5}  & 39.6  & 21.3  & 40.4  & 47.5 \\\bottomrule
\end{tabular}
\end{table}


\begin{table*}
\centering
\caption{Incremental results (\%) based on GFLV1 detector on COCO benchmark under last 40 classes + first 40 classes. (``$\Delta$'' represents an improvement over Catastrophic Forgetting. ``$\nabla$'' represents the gap towards the Upper Bound.)}
\label{table2}
\begin{tabular}{@{}l|cccccc@{}}
\toprule
Method & $AP$ & $AP_{50}$ & $AP_{75}$  & $AP_{S}$  & $AP_{M}$  & $AP_{L}$  \\ \midrule
Upper Bound & 40.2 & 58.3  & 43.6  & 23.2  & 44.1   & 52.2  \\
Catastrophic Forgetting & 22.6 & 32.7  & 24.2 & 15.1  & 25.0 & 27.6\\ \midrule
LwF \cite{DBLP:journals/pami/LiH18a} & 20.5 ($\Delta -2.1/\nabla 19.7$) & 29.9  & 22.1 & 13.0  & 22.5 & 25.3\\
RILOD \cite{DBLP:conf/edge/LiTGZZH19} & 34.1 ($\Delta 11.5/\nabla 6.1$) & 51.1  & 36.8 & 19.1  & 38.0 & 43.9\\
SID \cite{DBLP:journals/cviu/PengZMLL21} & 33.5 ($\Delta 10.9/\nabla 6.7$) & 50.9  & 36.3 & 19.0  & 37.7 & 43.0\\
ERD & \textbf{37.5 ($\Delta 14.9/\nabla 2.7$)} & \textbf{55.1}  & \textbf{40.4}  & \textbf{21.3}  & \textbf{41.1}   & \textbf{48.2}\\ \bottomrule
\end{tabular}
\vspace{-0.4em}
\end{table*}

\subsection{Ablation Study}

We validate the effectiveness of each component of the proposed method on MS COCO. In Table \ref{table3}, ``KD'' denotes only use the distillation loss without selection, while ``ERD'' denotes the selection strategy are introduced.  ``all cls + all reg'' denotes responses from both classification and regression branch are treated equally in the incremental process, which is used as our baseline. ``all cls'' denotes all classification responses in the incremental process are treated equally. ``all reg'' denotes all regression responses are treated equally in the incremental process. ``cls + ERS'' denotes that the ERS strategy is employed on the classification branch to conduct incremental distillation, as shown in Equation \ref{eq4}. ``cls + reg + ERS'' denotes responses on regression branch are added as well, as shown in Equation \ref{eq7}. In Table \ref{table3}, distillation on either classification or regression branch can merely obtain a low performance (i.e. 23.8\% and 13.0\% of AP). When all responses from the regression branch are used, AP is even lower than the fine-tuning strategy, which supports our findings shown in Figure \ref{figure0}. Comparatively, when combined responses from classification with regression branch, the AP reaches to 31.5\%. When ERS is involved to select responses from classification branch, the student detector can obtain higher results (i.e. 33.2\%). Furthermore, when performing ERS on regression branch, the AP continually increases to 36.9\%, which is a dramatically improvement (i.e. 5.4\%) compared with the baseline. All these results clearly point out the advantages of the proposed method.

\noindent \textbf{Parameter $\alpha$.} We conduct four groups of experiments to investigate the robustness of the proposed method on the parameter $\alpha$, which is utilized to elastically select positive responses from classification head and regression head. In table \ref{table5}, different combinations of $\alpha_{1}$ and $\alpha_{2}$ are chosen from the set ([1,1], [1,2], [2,1], [2,2]) to perform the training process. We observe that the maximum performance gap is merely 0.4\%, which indicates the proposed ERS is insensitive to the parameter $\alpha$. Therefore, the proposed ERS can be regarded as nearly parameter-free.

\begin{table}
\scriptsize
\centering
\caption{Incremental results (\%) based on FCOS detector.}
\label{table7}
\begin{tabular}{@{}l|l|c|c|cccc@{}}
\toprule
Model              & Method               & Centerness & Elastic & $AP$ & $AP_{50}$ & $AP_{75}$ \\ \midrule
                      & Upper Bound & \ding{52} &   & 38.5  & 57.5  & 41.3  \\
                      & Fine-tuning & \ding{52} &   & 16.7  & 25.6  & 17.9  \\ \cmidrule(l){2-7}
\multirow{2}{*}{FCOS} & \multirow{2}{*}{All} &         &            & 31.5  & 49.6  & 33.2 \\ 
                      &                      & \ding{52} &          & 31.7  & 49.9  & 33.3 \\ \cmidrule(l){2-7} 
                      & \multirow{2}{*}{ERD} &           & \ding{52}  & \textbf{34.4}  & \textbf{52.8}  & 36.5 \\  
                      &                      & \ding{52} & \ding{52}  & 34.2  & 52.4  & 36.6 \\ \bottomrule
\end{tabular}
\end{table}

\subsection{Discussions}
In this section, we present further insights into response-based IOD.

\noindent\textbf{Generalization on different detectors.} We perform extended experiments to validate the generality of the proposed method on the FCOS detector. For FCOS, we only need to replace the LD loss with GIoU loss. For both regression and centerness branches, we employ the statistical characteristics of category information to determine the elastic responses. Other settings are consistent with the proposed method. Results in Table \ref{table7} show that our method still brings stable gain regardless of the detector structure. To sum up, we only need to adjust our method slightly for adapting the head of different detectors, which indicates the generalizability of the proposed method.

\noindent\textbf{Elastic response helps both learning and generalization.} Considering the long-tail distribution of COCO, we configure an experiment under the last 40 classes + first 40 classes scenario. In this case, objects of the first 40 classes contain more information, which means more responses could be obtained. As shown in Table \ref{table2}, the performance can be further improved, with a smaller gap 2.7\% against the upper bound, which indicates the proposed method benefits from more responses to alleviate catastrophic forgetting.

\begin{table}
\scriptsize
\centering
\caption{Quantitative results (\%) of feature-based and response-based solutions.}
\label{table6}
\begin{tabular}{@{}l|c|c|c|cccccc@{}}
\toprule
Method & Feature & Response & Elastic & $AP$ & $AP_{50}$ & $AP_{75}$  \\ \midrule
All    &  & \ding{52} &   & 31.5 & 48.3  & 33.4 \\ 
FPN + All & \ding{52} & \ding{52} &   & 32.5  & 49.7  & 34.4 \\
FPN + ERS & \ding{52} & \ding{52} & \ding{52}  & 36.5  & 54.0  & 39.0 \\
ERD     &  & \ding{52} & \ding{52}  & \textbf{36.9} & \textbf{54.5}  & \textbf{39.6} \\\bottomrule
\end{tabular}
\end{table}

\begin{figure}
\centering
\includegraphics[width=5cm]{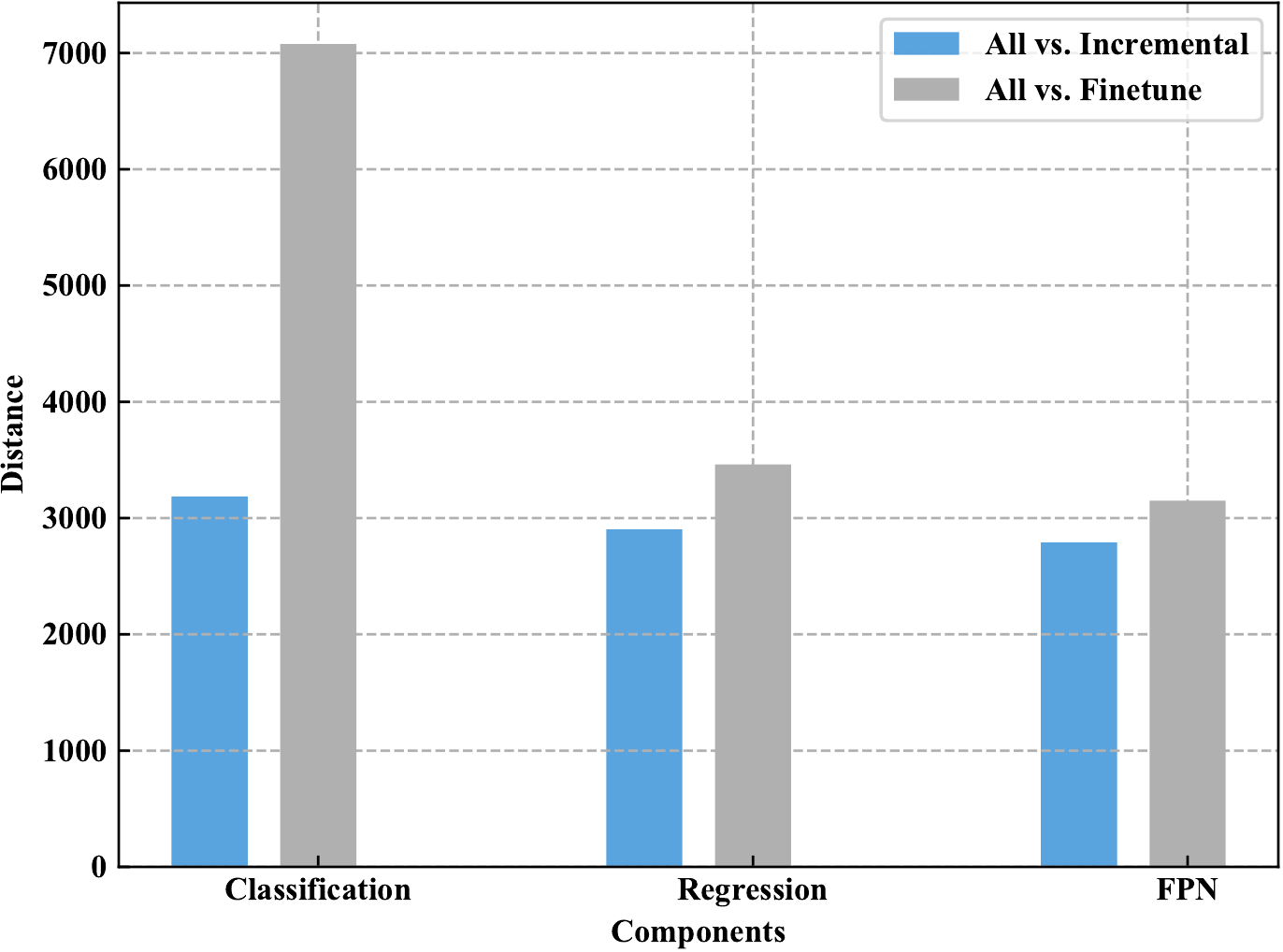}
\caption{Feature distance analysis of different components.}
\label{figure5}
\vspace{-1.8em}
\end{figure}

\noindent\textbf{Distances of different components.} In order to verify why the response-based distillation can attain higher performance compared to feature-based solutions, we randomly choose 10 images from COCO minival and calculate the L2 feature distances in varying components using different training strategies. As shown in Figure \ref{figure5}, ``All'' denotes the full data training strategy; ``Finetune'' denotes the fine-tuning strategy; ``Incremental'' denotes the proposed method. When compared ``All vs. Incremental'' with ``All vs. Finetune'', the distance of classification head is larger than that of regression head, and distances of the former two are larger than that of FPN (i.e. feature layers). It means that the response-based distillation provides more contributions to alleviate catastrophic forgetting.

\noindent\textbf{Quantitative analysis of feature-based and response-based solution.} Besides qualitative analysis in Figure \ref{figure5}, we further analyze the quantitative difference between feature-based and response-based solutions. As shown in table \ref{table6}, when combined FPN (i.e. feature layers) with all responses in head, it would produce positive effects. The reason is that feature layers provide more capacity for the learning procedure compared with head alone. Nevertheless, when the ERS strategy is added to head, the final performance is dramatically improved (32.5\% vs. 36.9\%), while the involvement of feature layers brings negative impacts (-0.4\% in AP). We guess a feasible explanation could be the optimization directions are changed, as feature layers tend to a global direction while head expects to reserve positive responses after selection.

\section{Conclusion}

In this paper, we elaborately design a response-based incremental paradigm in object detection field, which significantly alleviates the catastrophic forgetting problem. Firstly, we learn responses from the classification head and regression head, and specifically introduce incremental localization distillation in regression responses. Secondly, the elastic selection strategy is designed to provide suitable responses in different heads. Extensive experiments validate the effectiveness of the proposed method. Finally, elaborate analysis discusses the generalizability of our method and essential differences between response-based and feature-based distillation for incremental detection task, which provides insights for further exploration in this field.

\section*{Broader Impact}

The study of IOD would make us better understand the formation mechanism of neural networks from the system level, which provides a technical basis for the development of lifelong learning mechanism. The ultimate goal is that detectors can perform continual learning like creatures. However, models after incremental learning may lead to some privacy issues, while we can mitigate it by limiting the accessibility of trained models.

{\small
\bibliographystyle{ieee_fullname}
\bibliography{refs}
}

\clearpage
\section*{Supplementary materials}

\begin{figure*}[t]
\centering
\begin{subfigure}{.4\textwidth}
  \centering
  \includegraphics[width=.8\linewidth]{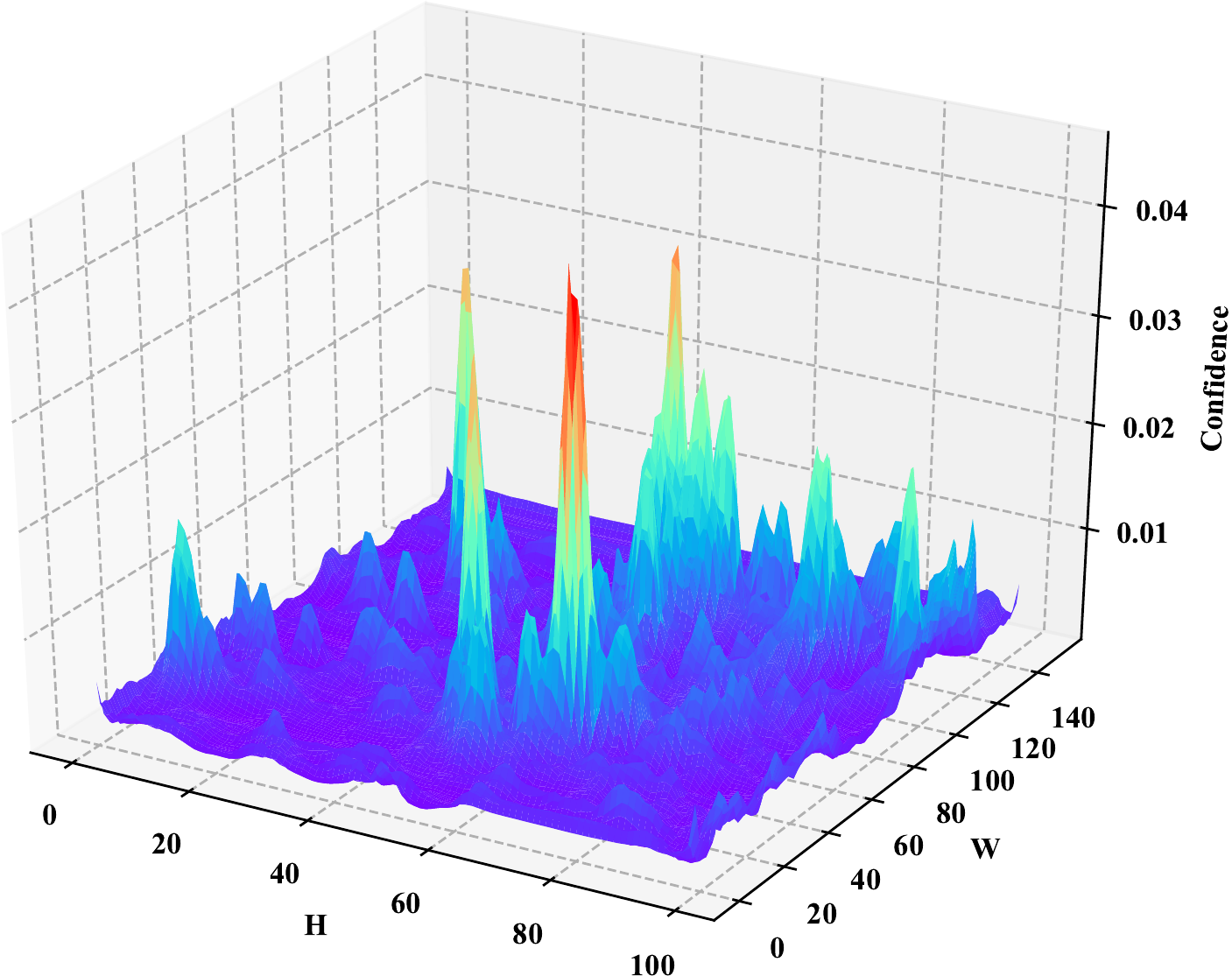}  
  \caption{Original classification responses on sample 1}
  \label{fig:sub-materials11}
\end{subfigure}
\begin{subfigure}{.4\textwidth}
  \centering
  \includegraphics[width=.8\linewidth]{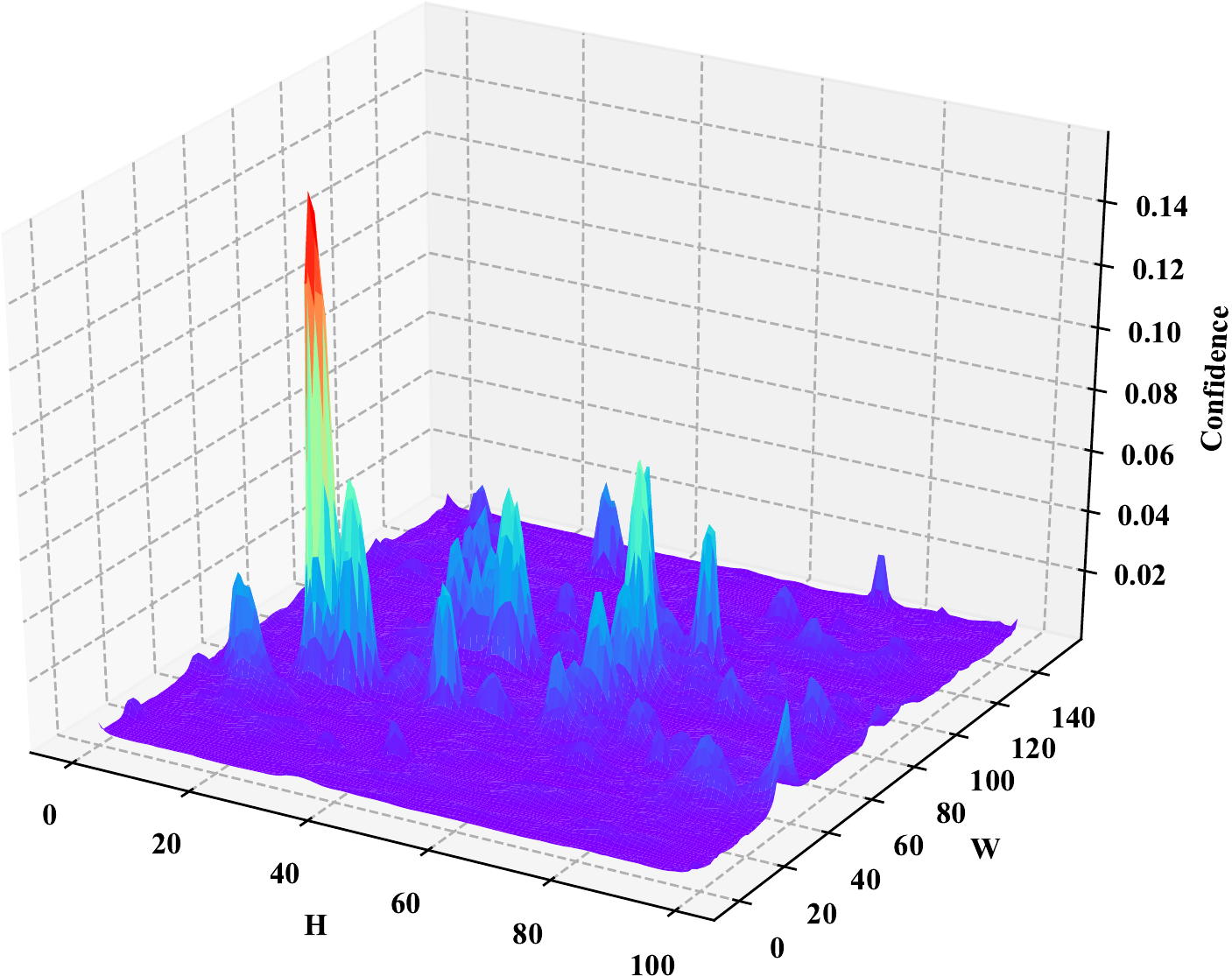}  
  \caption{Original classification responses on sample 2}
  \label{fig:sub-materials12}
\end{subfigure}
\begin{subfigure}{.4\textwidth}
  \centering
  \includegraphics[width=.8\linewidth]{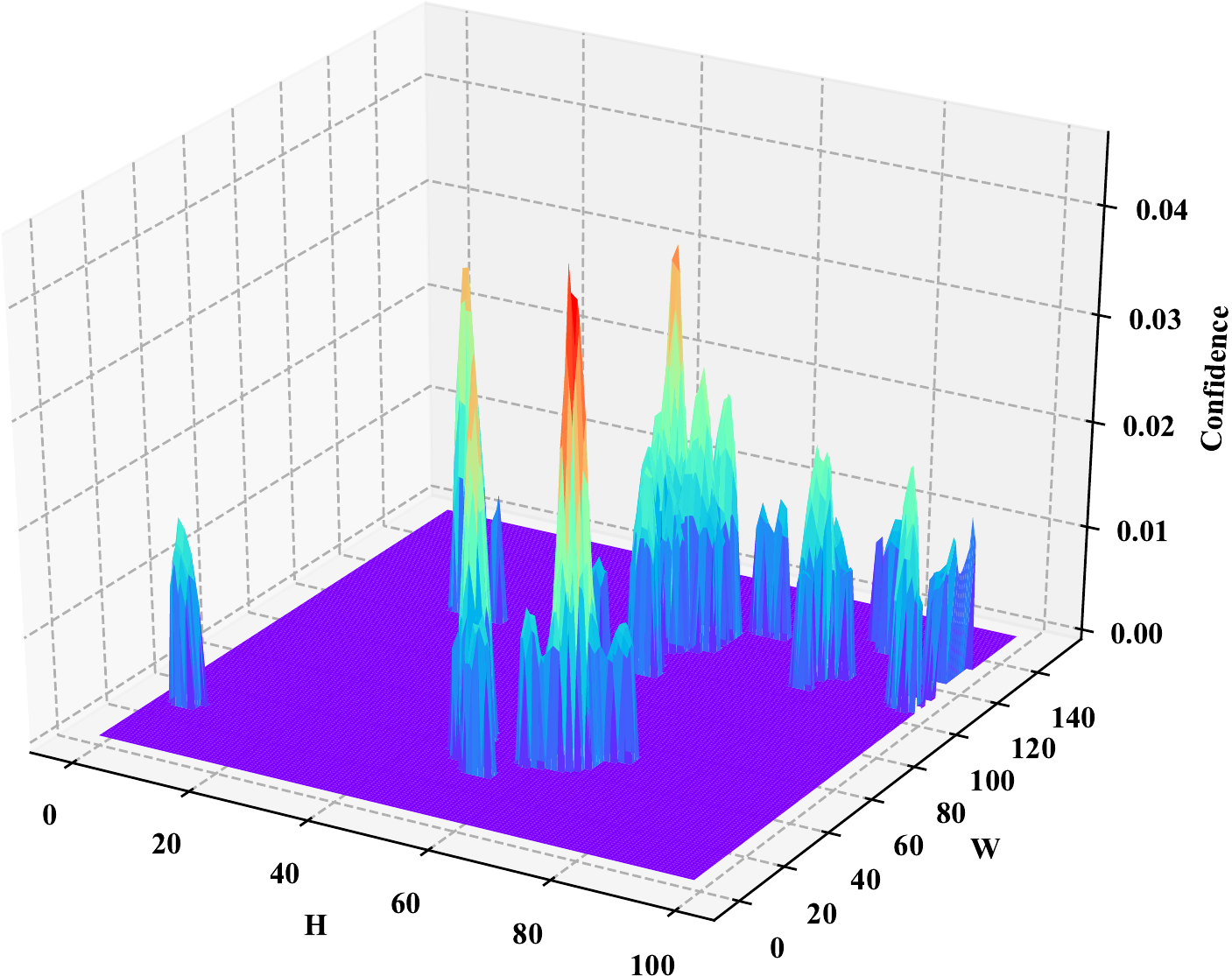}  
  \caption{Classification responses after using ERS on sample 1}
  \label{fig:sub-materials13}
\end{subfigure}
\begin{subfigure}{.4\textwidth}
  \centering
  \includegraphics[width=.8\linewidth]{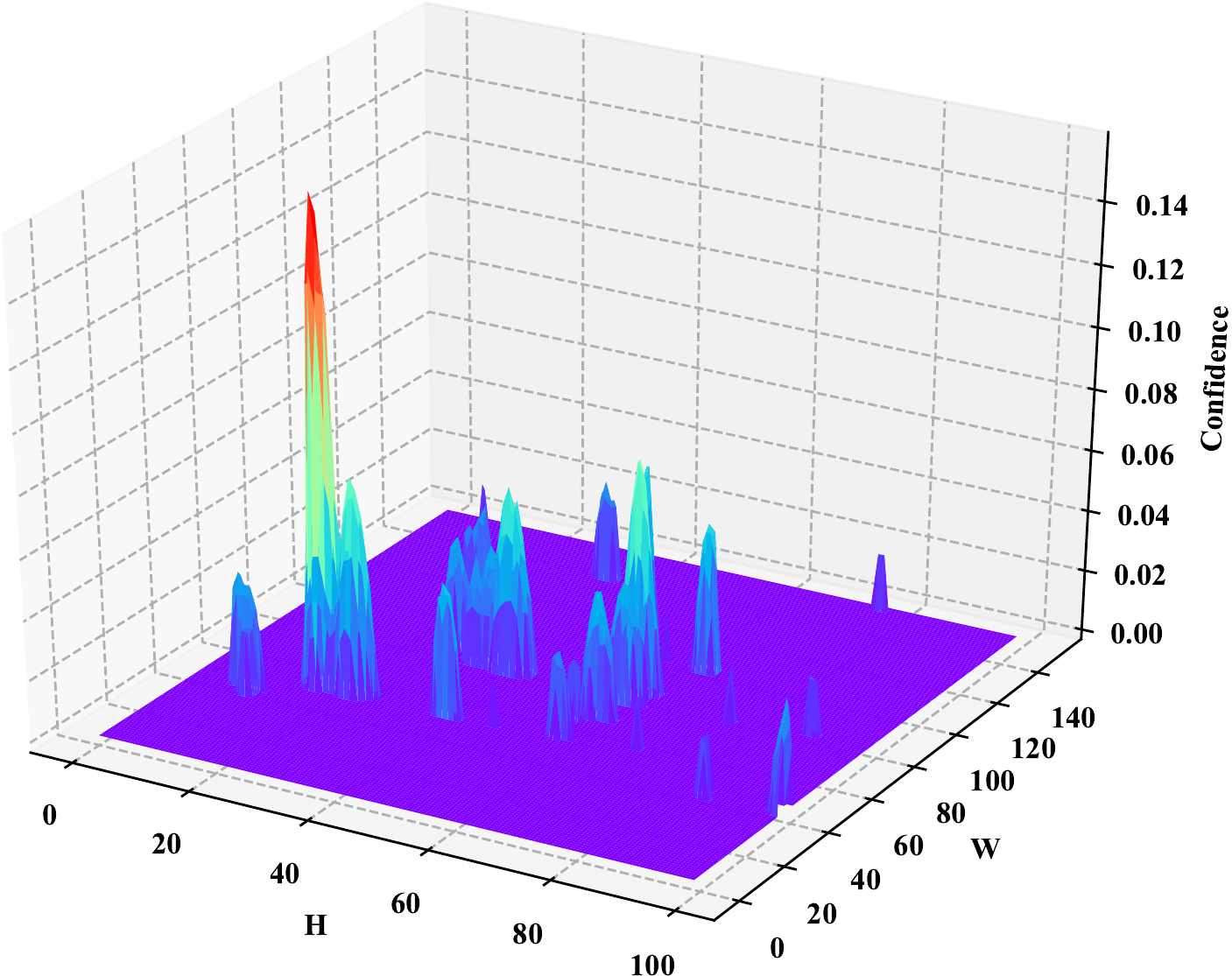}  
  \caption{Classification responses after using ERS on sample 2}
  \label{fig:sub-materials14}
\end{subfigure}
\caption{Visualization of responses in classification head (P3 level) between different samples.}
\label{fig:materials1}
\end{figure*}

\begin{figure*}
\centering
\begin{subfigure}{.4\textwidth}
  \centering
  \includegraphics[width=.8\linewidth]{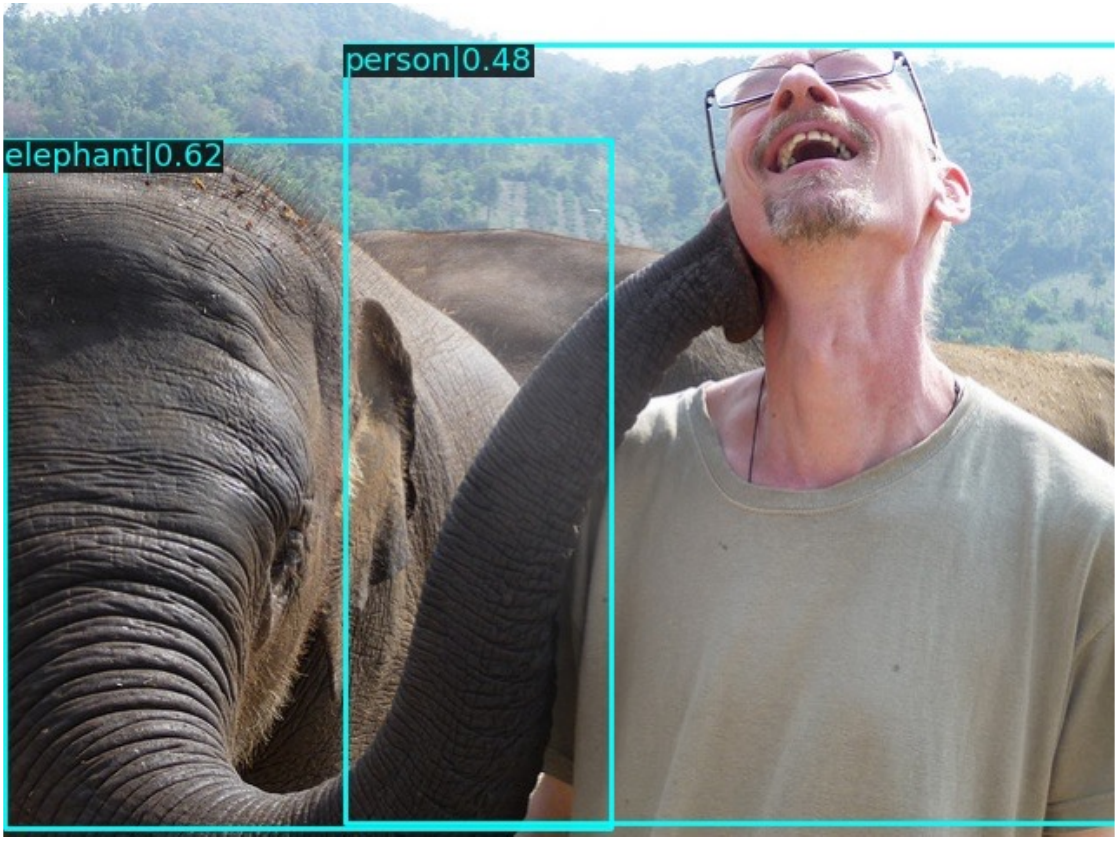}  
  \caption{w/o LD loss}
  \label{fig:sub-materials31}
\end{subfigure}
\begin{subfigure}{.4\textwidth}
  \centering
  \includegraphics[width=.8\linewidth]{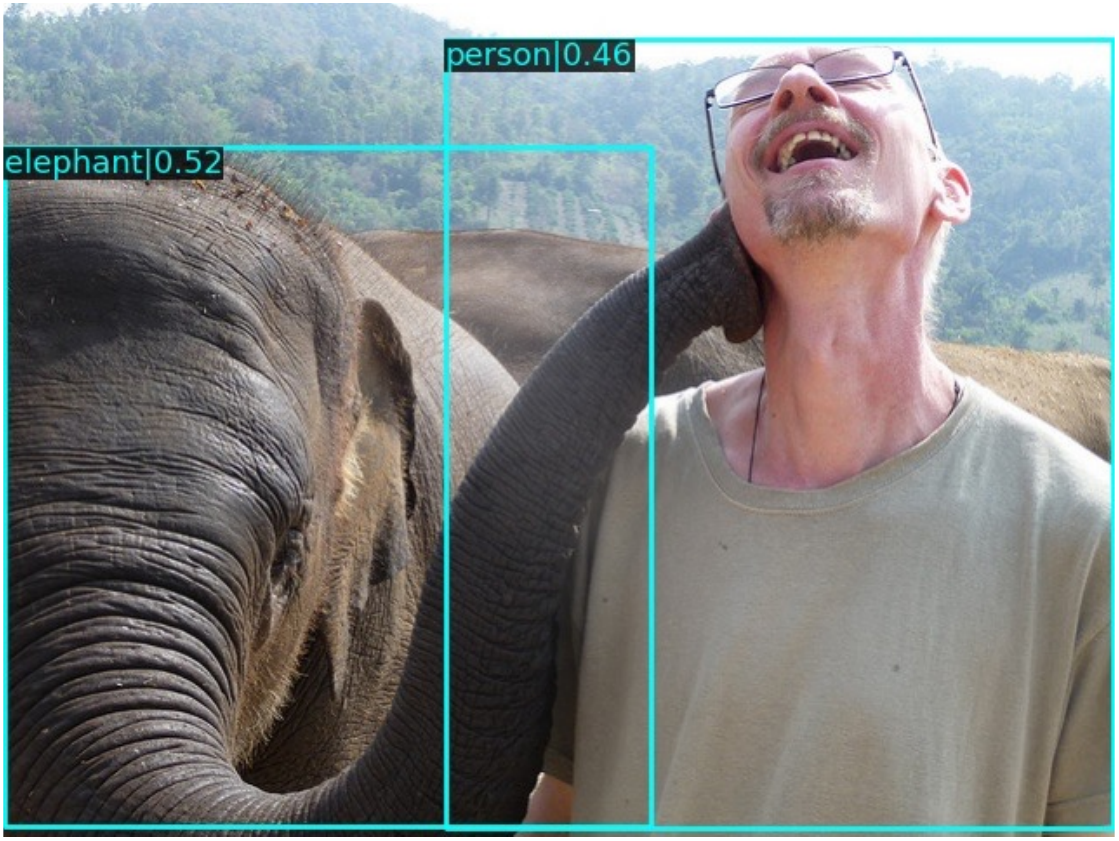}  
  \caption{w LD loss}
  \label{fig:sub-materials32}
\end{subfigure}
\caption{Detection results of w/o LD loss vs. w LD loss.}
\label{fig:materials3}
\end{figure*}

This section contains more experimental results and discussions.

\renewcommand\thesection{\Alph{section}}
\setcounter{section}{0}

\section{More discussions about responses of the head.}

We visualize the response in the classification head of the P3 level (the first layer of the FPN) between different samples in Figure \ref{fig:materials1}. Z-axis indicates the classification confidence score of each position. The outputs of classification head on P3 level is $100*152$, which indicates the number of all responses on P3 level. In Figure \ref{fig:sub-materials11} (sample 1) and Figure \ref{fig:sub-materials12} (sample 2), the positive responses (high-confidence) accounted for only a fraction of all responses, while the majority of the remaining responses are negative (low-confidence close to 0). This visually demonstrates that the response selection (ERS) is essential. Secondly, the maximum confidence for responses in sample 1 is approximately 0.04, while the maximum confidence for responses in sample 2 is approximately 0.14, which is 2.5 times higher than that in sample 1. It's a common situation that different images have enormously different peak response confidences. Selecting them by a fixed number will result in redundant background responses or lacking foreground responses among different images. As shown in Figure \ref{fig:sub-materials13} and Figure \ref{fig:sub-materials14}, if equipped with ERS, the model performs a fairer selection as stated in Section 3.5 of the main paper, which also supports our motivation for the proposed ERS.

\section{Detailed results of various responses for IOD.}

In the Introduction section, we present that not all responses are important to prevent catastrophic forgetting. In this subsection, we provide detailed results of Figure \ref{figure0} in Table \ref{sub-table3}. We gradually increase the number of responses from 5 to all, corresponding to the less responses to all responses of Figure \ref{figure0}. The tables indicates that although by exhaustively traversing the number of responses with appropriate step sizes can find an adequate global number, it still trails our elastic method.

\begin{table}[htbp]
\centering
\caption{Performance (\%) of various responses for IOD under first 40 classes + last 40 classes scenario.}
\label{sub-table3}
\begin{tabular}{@{}l|cccccc@{}}
\toprule
Response & $AP$ & $AP_{50}$ & $AP_{75}$ & $AP_{S}$  & $AP_{M}$  & $AP_{L}$  \\ \midrule
\# = 5    & 30.0 & 44.3  & 32.1 & 13.0 & 33.9 & 41.4 \\
\# = 10   & 32.3 & 47.3  & 34.7 & 14.9 & 36.4 & 43.6 \\ 
\# = 50   & 35.6 & 52.5  & 38.3 & 19.0 & 39.2 & 46.4 \\ 
\# = 100  & \textbf{36.3} & \textbf{53.4}  & \textbf{39.0} & 19.7 & 39.9 & 46.6 \\ 
\# = 1k   & 36.1 & 53.9  & 38.7 & 20.0 & 40.1 & 46.8 \\ 
\# = 2k   & 35.5 & 53.1  & 38.1 & 19.4 & 39.4 & 46.6 \\ 
\# = 4k   & 34.9 & 52.7  & 37.1 & 19.6 & 38.8 & 45.7 \\ 
\# = 6k   & 34.8 & 52.4  & 36.9 & 19.6 & 38.6 & 45.0 \\ 
\# = 8k   & 34.1 & 51.5  & 36.2 & 19.1 & 37.8 & 45.0 \\ 
\# = 10k  & 33.7 & 51.2  & 35.8 & 18.9 & 37.3 & 44.1 \\
\# = all  & 31.5 & 48.3  & 33.4 & 17.7 & 35.3 & 41.3 \\ \bottomrule
\end{tabular}
\end{table}

\section{Detailed results of base classes and new classes.}

In Table \ref{sub-table11}, we provide detailed results of base classes and new classes for Table \ref{table11}. The“Upper Bound” denotes results of base and new classes by performing a standard training with full data. “ERD” denotes results of base and new classes by performing training with our method in an incremental setting. In Table \ref{table11}, catastrophic forgetting leads to detection results of the base class collapsed to almost 0. As shown in Table \ref{sub-table11}, our method significantly improves the performance of the base classes at the cost of a slight decrease in the performance of the new classes.

\begin{table}[htbp]
\centering
\caption{Performance (\%) of base classes vs. new classes under different scenarios.}
\label{sub-table11}
\begin{tabular}{@{}l|l|l|c@{}}
\toprule
Scenarios & Method & Classes & $AP$  \\ \midrule
\multirow{2}{*}{Full data} & \multirow{2}{*}{Upper Bound} & Base  & 45.4 \\ 
                                                    &     & New   & 35.0  \\ \midrule
\multirow{2}{*}{40 classes + 40 classes}& \multirow{2}{*}{ERD}    & Base  & 41.6 \\ 
&                         & New   & 32.1 \\\bottomrule
\toprule
\multirow{2}{*}{Full data} & \multirow{2}{*}{Upper Bound} & Base  & 42.1\\ 
                                                    &     & New   & 37.0  \\ \midrule
\multirow{2}{*}{50 classes + 30 classes}& \multirow{2}{*}{ERD}    & Base  & 38.0 \\ 
&                         & New   & 34.3 \\\bottomrule
\toprule
\multirow{2}{*}{Full data} & \multirow{2}{*}{Upper Bound} & Base  & 40.9 \\ 
                                                    &     & New   & 38.2  \\ \midrule
\multirow{2}{*}{60 classes + 20 classes}& \multirow{2}{*}{ERD}    & Base  & 35.3 \\ 
&                         & New   & 37.1 \\\bottomrule
\toprule
\multirow{2}{*}{Full data} & \multirow{2}{*}{Upper Bound} & Base  & 41.6 \\ 
                                                    &     & New   & 30.4  \\ \midrule
\multirow{2}{*}{70 classes + 10 classes}& \multirow{2}{*}{ERD}    & Base  & 35.7 \\ 
&                         & New   & 28.8 \\\bottomrule
\end{tabular}
\end{table}

\section{Ablation study of LD loss.}

Considering that the LD distillation loss is limited for detectors other than GFL. Therefore, we use the more general L2 distillation loss instead of LD distillation loss to validate our method. As shown in Table \ref{sub-table2}, even though dropping LD loss causes the unavailability of extra localization information, our method still maintains a high-level performance (36.4 \%). This validates a strong generalization of our method. 

In addition, we visualize the detection results of w LD loss and w/o LD loss in incremental scenes in Figure \ref{fig:materials3}. By comparing both two figures, our method presents a more accurate detection box.

\begin{table}[htbp]
\small
\centering
\caption{Performance (\%) of w/o LD loss vs. w LD loss.}
\label{sub-table2}
\begin{tabular}{@{}l|cccccc@{}}
\toprule
Loss & $AP$ & $AP_{50}$ & $AP_{75}$ & $AP_{S}$  & $AP_{M}$  & $AP_{L}$  \\ \midrule
w/o LD & 36.4 & 54.1 & 38.8 & 20.6 & 40.0 & 47.3 \\ 
w LD   & \textbf{36.9} & \textbf{54.5}  & \textbf{39.6}  & \textbf{21.3}  & \textbf{40.4}   & \textbf{47.5} \\\bottomrule
\end{tabular}
\end{table}

\section{Additional Visualizations of ERD.}

\begin{figure*}[t]
\centering
\begin{subfigure}{.4\textwidth}
  \centering
  \includegraphics[width=.7\linewidth]{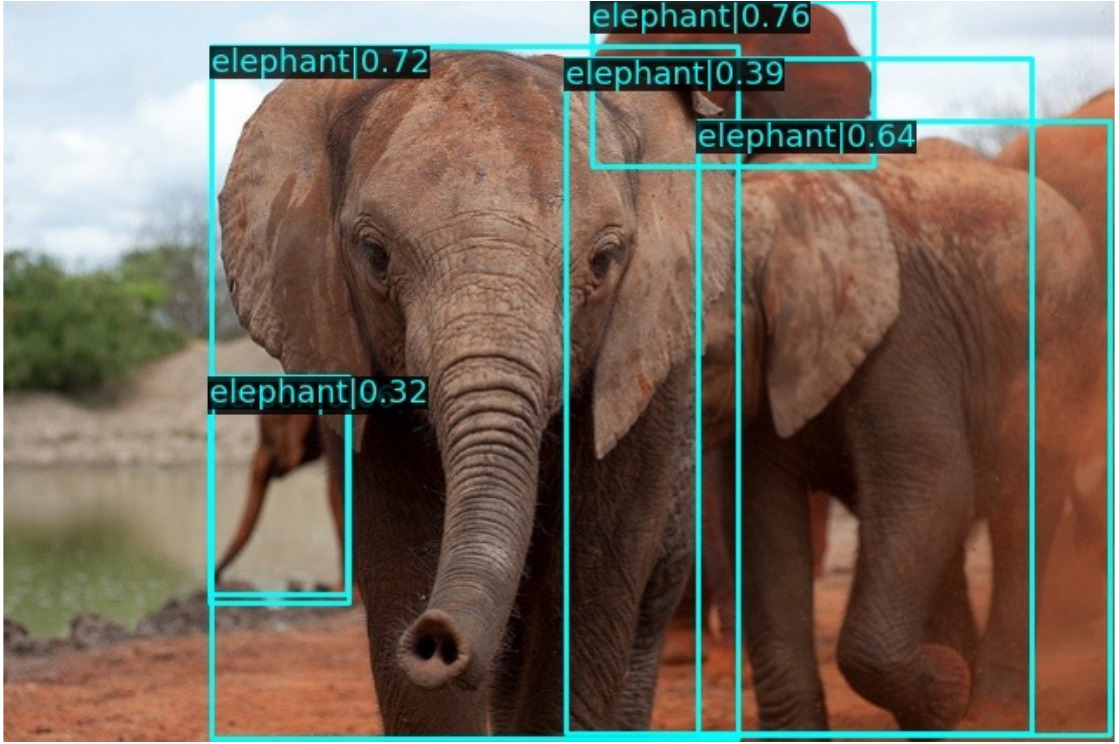}  
  \caption{Upper Bound}
  \label{fig:sub-materials21}
\end{subfigure}
\begin{subfigure}{.4\textwidth}
  \centering
  \includegraphics[width=.7\linewidth]{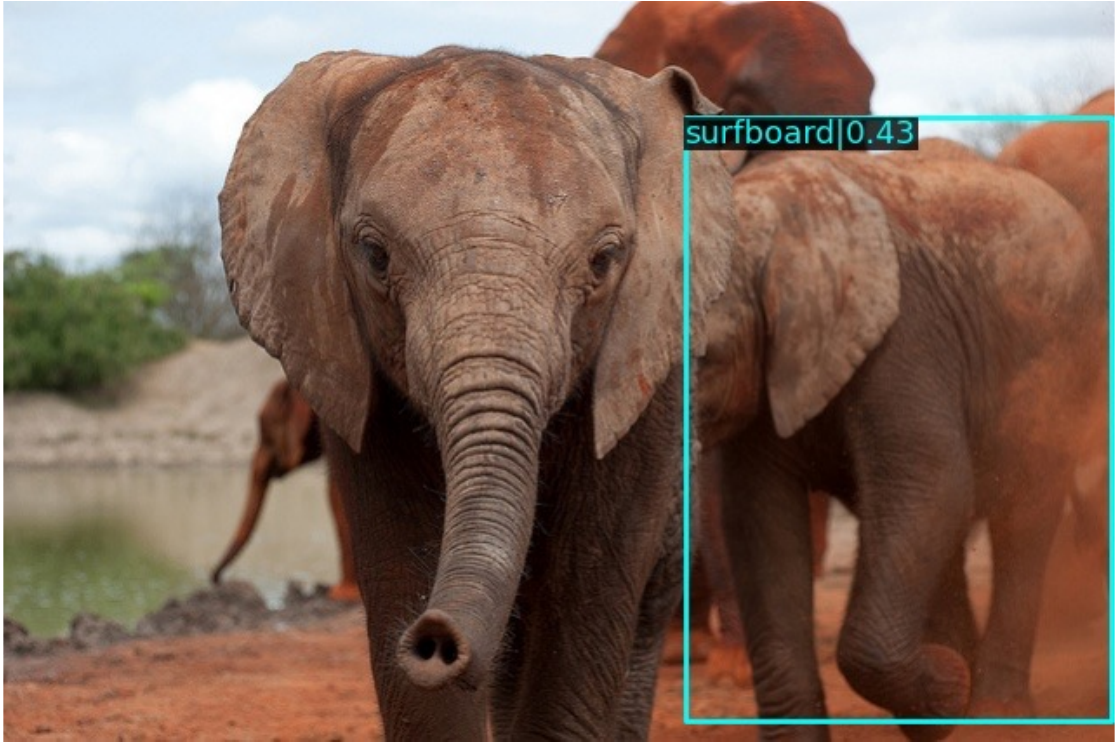}  
  \caption{Fine-tuning}
  \label{fig:sub-materials22}
\end{subfigure}
\begin{subfigure}{.4\textwidth}
  \centering
  \includegraphics[width=.7\linewidth]{show_result_SID_crop.pdf}  
  \caption{SID}
  \label{fig:sub-materials23}
\end{subfigure}
\begin{subfigure}{.4\textwidth}
  \centering
  \includegraphics[width=.7\linewidth]{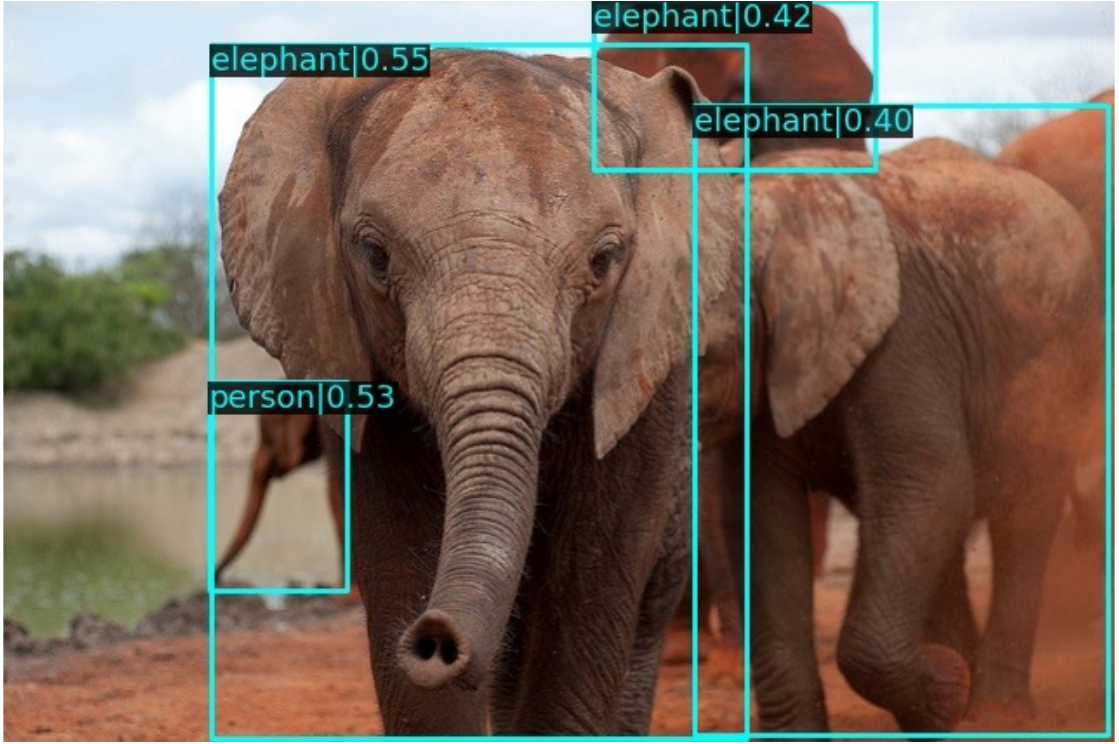}  
  \caption{ERD}
  \label{fig:sub-materials24}
\end{subfigure}
\caption{Visualization of experimental results before and after incremental learning: (a) Detection results with GFLV1 that train with all data. (b) Detection results with GFLV1 that fine-tune with new classes (Catastrophic Forgetting). (c) Incremental results of SID with GFLV1. (d) Incremental results of our method with GFLV1.}
\label{fig:materials2}
\end{figure*}

Figure \ref{fig:materials2} illustrates one example detected with different
schemes under first 40 classes + last 40 classes scenario. The detection results of our method (Figure \ref{fig:sub-materials24}) has a significant improvement over directly fine-tuning (Figure \ref{fig:sub-materials22}) and SID (Figure \ref{fig:sub-materials23}). Directly fine-tuning barely detects the objects due to catastrophic forgetting. SID also misses most of the objects, even though the AP reaches 34\%. Our method performs most closely to the Upper Bound method.

\begin{algorithm}
\caption{\textbf{E}lastic \textbf{R}esponse \textbf{S}election (ERS)}
\label{algorithm1}
\hspace*{0.02in} {\bf Input:}
Unlabeled image \bm{$I$}, teacher detector \bm{$\theta^{\prime}$} \\
\hspace*{0.02in} {\bf Output:}
Sampled response sets \bm{$\mathcal{C}$}, \bm{$\mathcal{B}$}
\begin{algorithmic}[1]
\State Inference \bm{$I$} with \bm{$\theta^{\prime}$} yields the classification score \bm{$\mathcal{C}^{\prime}$} and predicted distribution \bm{$\mathcal{B}^{\prime}$} \\

\Statex \emph{// Classification branch}
\For{$i=1$ to $\mathcal{C}^{\prime}$} 
    \State $G_{\mathcal{C}^{\prime}}$ $\longleftarrow$ ${confidence(\mathcal{C}^{\prime}_{i})}$
\EndFor
\State Compute $\mu_{\mathcal{C}^{\prime}}=mean(G_{\mathcal{C}^{\prime}})$
\State Compute $\sigma_{\mathcal{C}^{\prime}}=std(G_{\mathcal{C}^{\prime}})$
\State Compute threshold $\tau_{\mathcal{C}^{\prime}}=\mu_{\mathcal{C}^{\prime}} + \alpha_{1} \sigma_{\mathcal{C}^{\prime}}$
\For{each candidate $c$ in $\mathcal{C}^{\prime}$}
    \If{$G_{\mathcal{C}^{\prime}_i} \geq \tau_{\mathcal{C}^{\prime}}$}
    \State Add candidate $c$ to \bm{$\mathcal{C}$}
    \EndIf
\EndFor
\State \Return \bm{$\mathcal{C}$} \\

\Statex \emph{// Regression branch}
\For{$j=1$ to $\mathcal{B}^{\prime}$} 
    \State $G_{\mathcal{B}^{\prime}}$ $\longleftarrow$ ${\operatorname{Top-1}(\mathcal{B}^{\prime}_{j})}$
\EndFor
\State Compute $\mu_{\mathcal{B}^{\prime}}=mean(G_{\mathcal{B}^{\prime}})$
\State Compute $\sigma_{\mathcal{B}^{\prime}}=std(G_{\mathcal{B}^{\prime}})$
\State Compute threshold $\tau_{\mathcal{B}^{\prime}}=\mu_{\mathcal{B}^{\prime}} + \alpha_{2} \sigma_{\mathcal{B}^{\prime}}$
\For{each candidate $b$ in $\mathcal{B}^{\prime}$}
    \If{$G_{B^{\prime}_j} \geq \tau_{\mathcal{B}^{\prime}}$}
    \State Add candidate $b$ to \bm{$\mathcal{B}$}
    \EndIf
\EndFor
\State \bm{$\mathcal{B}$} $\longleftarrow$ $nms(\mathcal{B},G_{\mathcal{B}})$

\State \Return \bm{$\mathcal{B}$}
\end{algorithmic}
\end{algorithm}

\end{document}